\documentclass{article} 
\usepackage{iclr2027_conference,times}

\usepackage{amsmath,amsfonts,bm}

\def\eqref#1{equation~\ref{#1}}

\def\1{\bm{1}}

\DeclareMathAlphabet{\mathsfit}{\encodingdefault}{\sfdefault}{m}{sl}
\SetMathAlphabet{\mathsfit}{bold}{\encodingdefault}{\sfdefault}{bx}{n}

\usepackage[hidelinks]{hyperref}
\usepackage{url}
\usepackage{wrapfig}
\usepackage{caption}
\usepackage{threeparttable}
\usepackage{multirow}
\usepackage{graphicx}
\usepackage{float}
\usepackage{placeins}
\usepackage{booktabs}
\usepackage{fontawesome5}
\usepackage{pifont}
\usepackage{xcolor}
\usepackage{twemojis}
\usepackage[T1]{fontenc}

\newdimen\logobase
\newdimen\logoht
\newdimen\logoraise
\title{\noindent
\setbox1=\hbox{Entity3R: An End-to-End Entity-Aware}%
\setbox2=\hbox{Geometric Foundation Model}%
\logobase=\dimexpr\ht1+\dp2+\baselineskip\relax
\logoht=1.15\logobase
\logoraise=\dimexpr-\baselineskip-\dp2-(\logoht-\logobase)/2\relax
\setbox0=\vtop{%
  \lineskiplimit=-\maxdimen
  \hbox{\unhbox1}%
  \hbox{\unhbox2}%
}%
\raisebox{\logoraise}[\ht0][\dp0]{%
  \includegraphics[trim=8 53 8 52,clip,height=\logoht]{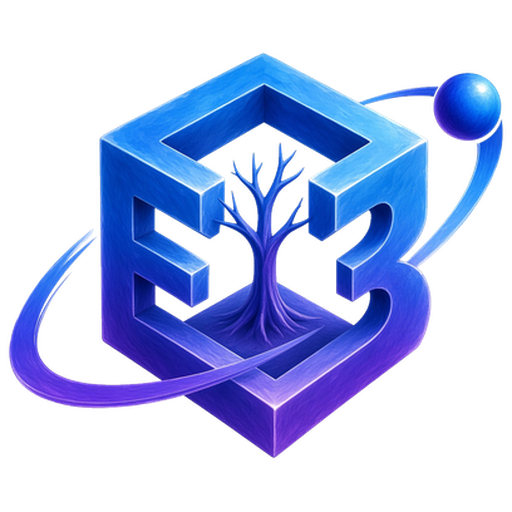}%
}\hspace{0.35em}\box0}

\author{%
  \textbf{Jinfeng Liu}$^{1,3}$ \quad
  \textbf{Runsong Zhu}$^{2,3}$ \quad
  \textbf{Wei Yin}$^{3}$ \quad
  \textbf{Xiaoyang Guo}$^{3}$ \quad
  \textbf{Yanqi Bao}$^{3}$ \\
  \textbf{Weiqiang Ren}$^{3}$ \quad
  \textbf{Qian Zhang}$^{3}$ \quad
  \textbf{Zhengqing Chen}$^{3}$\textsuperscript{\faEnvelope[regular]} \quad
  \textbf{Dan Xu}$^{1}$\textsuperscript{\faEnvelope[regular]} \\ \\
  $^1$The Hong Kong University of Science and Technology (HKUST) \\ $^{2}$The Chinese University of Hong Kong (CUHK) \quad $^{3}$Horizon Robotics \\
  \href{https://liujf1226.github.io/Entity3R}{\textcolor{magenta}{https://liujf1226.github.io/Entity3R}}
}

\newcommand{\B}[1]{\textbf{#1}}
\newcommand{\U}[1]{\underline{#1}}
\iclrfinalcopy 
\begin{document}

\maketitle

\begin{figure}[H]
\centering
\vspace{-5mm}
\includegraphics[width=\linewidth]{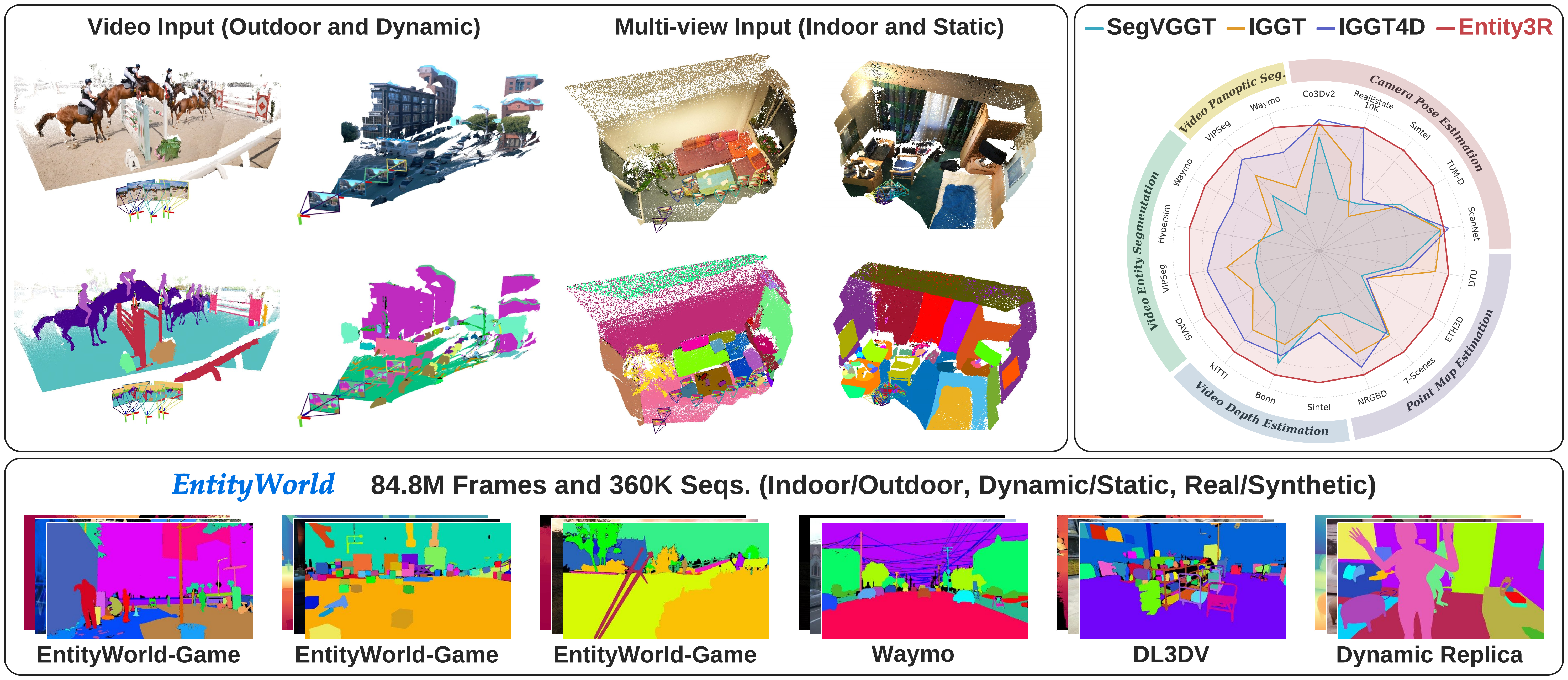}
\vspace{-7mm}
\caption{We present \textbf{Entity3R}, a unified end-to-end framework which can jointly perceive class-agnostic 3D entities in the open world and reconstruct scene geometry from several scene types. To address the data scarcity, we construct \textbf{EntityWorld}, an open-domain training corpus of nearly 84.8M frames and 360K sequences with geometry and entity masklet annotations, assembled from public geometric data resources and our collected game sequences. Entity3R achieves state-of-the-art performance on entity segmentation and geometric reconstruction benchmarks.}
\label{fig:teaser}
\end{figure}

\begin{abstract}
Class-agnostic 3D entity understanding aims to recover open-world entity identities with basic scene geometry from multi-view images or videos.
Recent state-of-the-art methods built on instance-grounded geometric foundation models rely on hyperparameter-sensitive clustering to decode objects, which makes entity mask inference unreliable and costly. They also inherit the low native resolution of geometry transformers, limiting fine-grained entity perception.
We present \textbf{Entity3R}, an end-to-end, entity-aware 3D/4D geometric foundation model. 
Entity3R extracts high-resolution geometry-instance representations with a dual-branch backbone, and decodes 3D/4D-consistent entity masks through learnable global object queries that interact with frame-level instance features.
This design enables accurate and efficient 3D entity segmentation without post-hoc clustering.
We further introduce a training-free, chunk-wise decoding strategy that allows Entity3R to decode more entities than its query budget and to process longer input sequences. 
To train Entity3R, we build \textbf{EntityWorld}, a large-scale corpus for joint geometry and instance learning, containing 360K sequences and 84.8M frames drawn from existing geometric datasets and newly collected photorealistic game data. Entity masklets are generated at scale by an automatic annotation pipeline.
Trained on EntityWorld, Entity3R achieves state-of-the-art performance on both entity segmentation and geometric reconstruction benchmarks.

\end{abstract}

\section{Introduction}
\label{sec:intro}

Spatial intelligence underpins embodied robotics and autonomous driving that must perceive their surroundings in order to navigate, manipulate objects, and interact physically.
One of its core capabilities is 3D entity understanding, which requires recovering the basic geometry of a scene and perceiving the entities that occupy it beyond the predefined categories.
This paper focuses on the task of class-agnostic 3D entity segmentation from the most widely used visual input (i.e., 2D images).
%
%

Earlier methods~\citep{bhalgat2023contrastive,zhu2025rethinking} mainly adopted an optimization-based multi-view fusion pipeline to lift 2D class-agnostic instance segmentation predictions to 3D scenes represented by reconstructed NeRF~\citep{mildenhall2021nerf} or 3DGS~\citep{threedgs}.
Although promising results have been achieved, these methods require time-consuming optimization (e.g., more than 10 minutes), and cross-view inconsistencies in 2D instance segmentation predictions may further propagate to the 3D lifting results.
More recently, the feedforward paradigm has revolutionized the field by designing class-agnostic instance-grounded geometric foundation models~\citep{iggt,iggt4d}.
Specifically, these models directly predict instance embeddings alongside reconstructed geometry from multi-view or video inputs.
Although improvements have been achieved by these methods, they are still limited by two limitations.
First, these methods require hyperparameter-sensitive clustering post-processing for object decoding, leading to unreliable and costly entity mask inference.
Second, they inherit the native resolution of current geometry transformers (518 pixels in width), which hinders the fine-grained segmentation for complicated scenes.


In this paper, we present \textbf{Entity3R}, a new unified end-to-end framework for effective class-agnostic entity understanding with geometric reconstruction, given multi-view images of static scenes or video frames of dynamic scenes.
We explicitly impose entity-level perception in Entity3R by introducing global object queries coupled with frame-level instance maps, in addition to the basic geometric properties (i.e., camera poses and depth maps).
In particular, we propose to learn fine-grained geometry-instance representation from high-resolution inputs with an efficient dual-branch design to preserve fine-grained entity information.
The object queries are then decoded into cross-view entity masks by attending to the resulted high-resolution instance features, which forms the global scene context.
%
During training, we match the mask predictions to ground-truth entities with the Hungarian algorithm~\citep{hungarian}, and each matched query is trained to predict the corresponding entity masklet. In this way, the queries and instance embeddings can be optimized via entity mask supervision, while the geometric properties are jointly learned with standard supervision following common practice~\citep{vggt}.
During inference, Entity3R decodes the queries into entity predictions \textit{in an end-to-end manner, improving accuracy over existing baselines while enabling superior efficiency} (e.g., $\mathbf{20}\times$ faster than the streaming clustering of IGGT4D~\citep{iggt4d}).
Moreover, we introduce a chunk-wise object decoding strategy to further boost inference scalability in the number of frames without any retraining, allowing Entity3R to \textit{cover more entities than a fixed set of queries can represent} in complex scenes and to \textit{support longer input sequences} (e.g., from 80 to 180 frames on a 24-GB consumer GPU).
%


In addition to the architectural innovation, we further contribute \textbf{EntityWorld}, a new dataset for unified 3D entity perception and geometric reconstruction.
To the best of our knowledge, \textit{EntityWorld is the largest dataset for joint geometry and instance learning}, covering both static and dynamic scenes, as well as both indoor and outdoor environments.
Besides existing public geometry datasets, EntityWorld includes a newly collected corpus of photorealistic game data with accurate geometry annotations, bringing the total to nearly 84.8M frames and 360K sequences.
To obtain 3D/4D-consistent entity masks, we design an automatic annotation pipeline.
Concretely, we first obtain initial entity mask proposals on the first frame with CropFormer~\citep{cropformer} and propagating them temporally with SAM2~\citep{sam2}.
An optical flow model~\citep{raft} then associates re-detections in uncovered regions with lost SAM2 tracks or newly appearing objects, and flags large propagation jumps at which we split the sequence into new clips.
The whole pipeline runs offline and automatically yields mask annotations at scale.

Our contributions are summarized as follows:
\begin{itemize}
\item We present \textbf{Entity3R}, a unified end-to-end framework for class-agnostic entity understanding with geometric reconstruction from multi-view images or video frames. It learns fine-grained geometry-instance representations from high-resolution inputs via an efficient dual-branch design, and jointly predicts cameras and depths while decoding 3D/4D-consistent entity masks by letting global object queries interact with instance features.
\item We introduce a chunk-wise object decoding strategy that extends inference scalability without retraining, allowing Entity3R to cover more entities than a fixed query set can handle in one pass and to support longer input sequences.
\item We construct EntityWorld, an open-domain training corpus of nearly 84.8M frames and 360K sequences with geometry and entity masklet annotations, spanning existing geometric datasets and our newly collected game data. Our Entity3R achieves state-of-the-art performance on entity segmentation and geometric reconstruction benchmarks.
\end{itemize}

\section{Related Work}
\label{sec:related}

\begin{figure}[t]
\centering
\includegraphics[width=\linewidth]{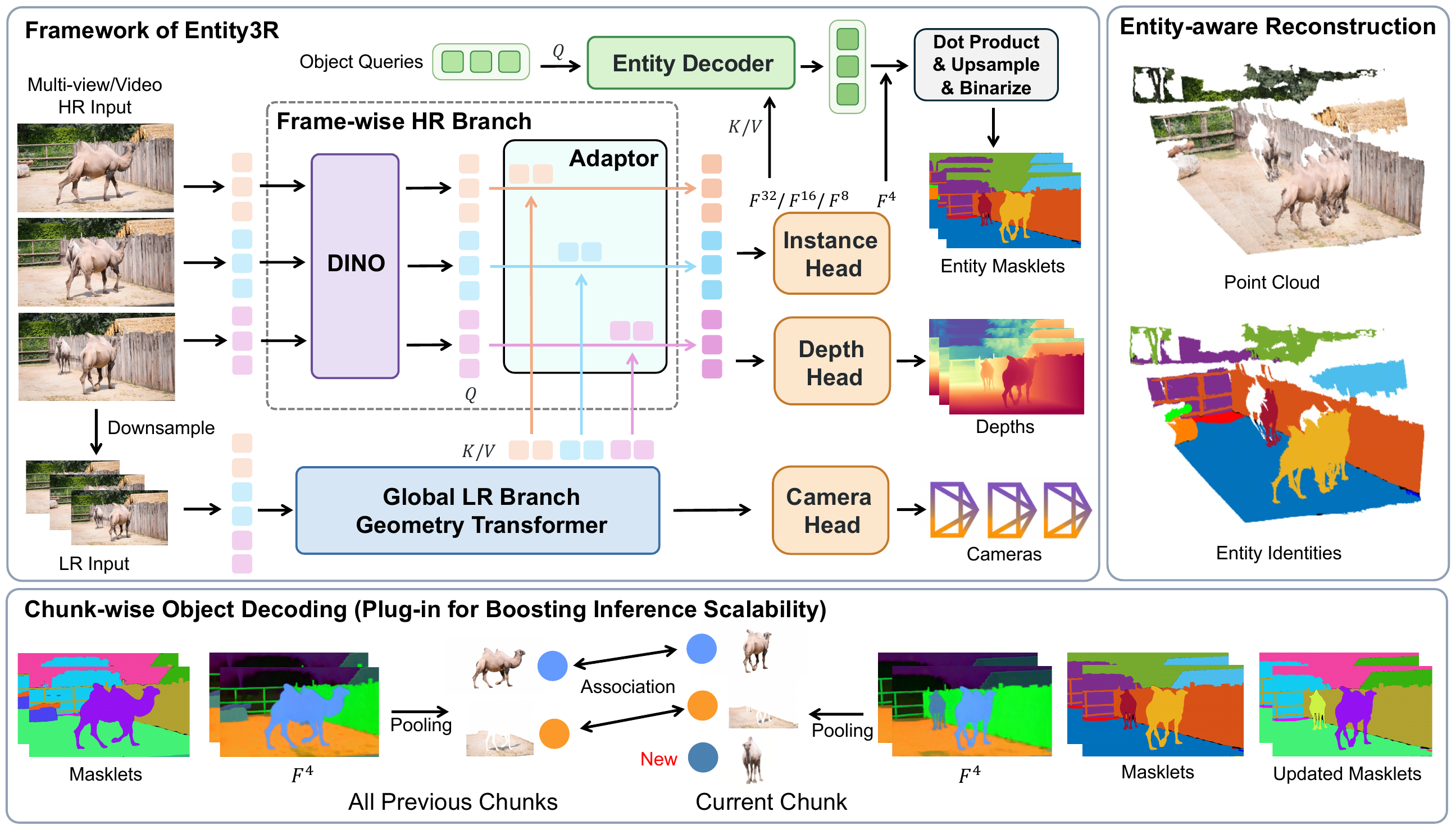}
\vspace{-7mm}
\caption{\textbf{Overview of Entity3R.} We achieve 3D class-agnostic entity understanding with geometric reconstruction in an end-to-end manner. Fine-grained geometry-instance representation is first extracted from high-resolution inputs using a dual-branch backbone. Then, the representation is fed into dense prediction heads for high-resolution depth and instance maps. Learnable object queries are updated in the entity decoder by alternately attending to the cross-view consistent instance features, and are decoded into class-agnostic entity masklets. We also propose a chunk-wise object decoding strategy as a memory-efficient plug-in for boosting inference scalability.}\vspace{-3mm}
\label{fig:overview}
\end{figure}

\noindent\textbf{Geometric Foundation Models.}~Feed-forward geometric foundation models~\citep{dust3r,mast3r,vggt,pi3,da3,vggtomega} reconstruct cameras and dense geometry from unposed multi-view or video inputs in a single forward pass, which is much more efficient than traditional SfM pipelines such as COLMAP~\citep{colmap}. DUSt3R~\citep{dust3r} regresses dense point maps from image pairs. Extending the reconstruction beyond two views then requires aligning those pairwise reconstructions after inference, which is costly and can be unstable. VGGT~\citep{vggt} removes this pairwise fusion and jointly predicts cameras, depths, and point maps for many views in one pass, with accuracy competitive with optimization-based pipelines. Subsequent geometry transformers follow this design: $\pi^3$~\citep{pi3} predicts permutation-equivariant cameras and point maps without a designated reference view, DA3~\citep{da3} predicts a singular depth-ray target with a plain transformer, and VGGT-$\Omega$~\citep{vggtomega} utilizes scales the training corpus with efficient register attention. Another line of work accelerates these transformers for long sequences~\citep{fastvggt,flashvggt,zipmap} and enables streaming reconstruction~\citep{cut3r,streamvggt,infinitevggt,lingbotmap}. All above models remain focused on geometric reconstruction and do not assign object identity to the predicted point clouds, whereas Entity3R predicts open-world entity identities attached to the recovered 3D/4D geometry.

\noindent\textbf{3D Scene Understanding.}~Integrating semantics into 3D reconstruction enables high-level reasoning and understanding of a scene. Optimization-based methods~\citep{langsplat,feature3dgs,onlinepg} inject vision-language features into 3D Gaussian Splatting~\citep{threedgs}. Feed-forward successors~\citep{lsm,uni3r,eps3d} predict semantic 3D Gaussians from unposed multi-view images in a single pass, without per-scene optimization. Most of these works follow a two-stage design. They first extract per-view semantic features with 2D VLMs~\citep{clip,lseg} and then fuse these features across views in 3D, which can accumulate cross-view errors. Instance-grounded geometry transformers instead reconstruct geometry and object identity in one stage, learning directly from cross-view consistent instance masks. Among them, SegVGGT~\citep{segvggt} trains 3D reconstruction and instance segmentation with semantic categories on ScanNet~\citep{scannet}, and therefore does not transfer beyond this closed-set indoor setting. PanSt3R~\citep{zust2025panst3r} predicts multi-view-consistent panoptic masks for static, unposed images by attaching a shared-query decoder to frozen DINOv2 and MUSt3R backbones, leaving the geometry head at its pretrained MUSt3R weights and merging masks with a QUBO post-process. IGGT~\citep{iggt} learns class-agnostic instance features through contrastive learning and obtains cross-view masklets by costly global clustering. Clustered masks provide instance identity rather than semantic features, and open-vocabulary understanding is executed by querying VLMs or MLLMs on these masks. Concurrent with our work, IGGT4D~\citep{iggt4d} replaces global clustering with a streaming procedure and extends training to dynamic indoor and driving data, yet object decoding still depends on scene-dependent clustering hyperparameters and can over- or under-segment. In contrast, Entity3R trains object queries~\citep{detr} with Hungarian matching~\citep{hungarian} and decodes entity masklets in an end-to-end and class-agnostic manner, using object queries without clustering.
\section{Methodology}
\label{sec:method}

\label{sec:entityworld}
\begin{figure}[t]
    \centering
    \includegraphics[width=1.0\textwidth]{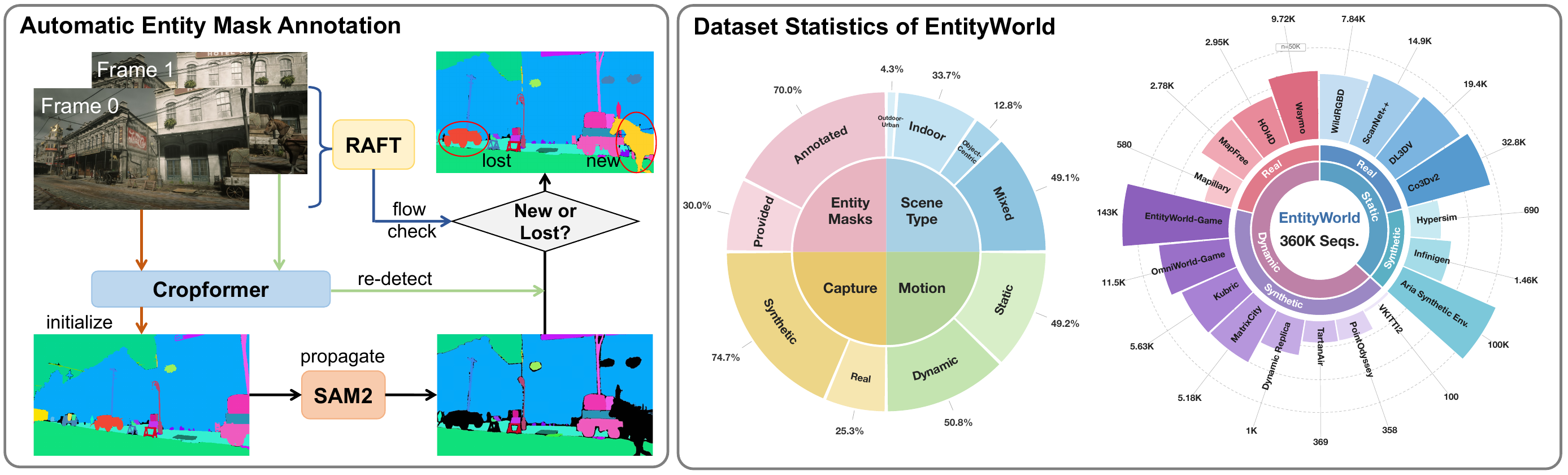}
    \vspace{-7mm}
    \caption{\textbf{Left:}~Our automatic entity mask annotation pipeline for scalable entity masklet generation. \textbf{Right:}~The dataset statistics of EntityWorld, which spanning diverse scene types.}
    \label{fig:entityworld}
    \vspace{-4mm}
\end{figure}

Entity3R is a class-agnostic 3D entity segmentation model that jointly reconstructs 3D/4D scene geometry and predicts open-world entity identities from unposed multi-view or video inputs. The overall architecture is illustrated in Fig.~\ref{fig:overview}. In the following, Sec.~\ref{sec:preliminaries} provides the preliminaries. Sec.~\ref{sec:backbone} describes our fine-grained geometry-instance representation. Sec.~\ref{sec:prediction} presents the end-to-end entity prediction. Sec.~\ref{sec:obj} gives the model training details.

\subsection{Preliminaries}
\label{sec:preliminaries}
\noindent\textbf{Visual Geometry Grounded Transformer.}~Given $N$ unposed RGB images $\{I_{i}\}_{i=1}^{N}$, where each $I_{i}\in\mathbb{R}^{3\times H\times W}$ comes from a multi-view image set or a video stream, VGGT~\citep{vggt} predicts cameras, depth maps and point maps in a single feed-forward pass: $f_{\theta}\big(\{I_{i}\}_{i=1}^{N}\big)=\big\{(\mathbf{g}_{i},D_{i},P_{i})\}_{i=1}^{N}$. Here $D_{i}\in\mathbb{R}^{H\times W}$ is the depth map of image $I_i$, $P_{i}\in\mathbb{R}^{3\times H\times W}$ is the global point map, and $\mathbf{g}_{i}=(\mathbf{q}_{i},\mathbf{t}_{i},\mathbf{f}_{i})\in\mathbb{R}^{9}$ is the concatenation of the rotation quaternion $\mathbf{q}_{i}\in\mathbb{R}^{4}$, the translation $\mathbf{t}_{i}\in\mathbb{R}^{3}$, and the field of view $\mathbf{f}_{i}\in\mathbb{R}^{2}$ parameterizing the camera of $I_i$. The geometry transformer $f_{\theta}$ consists of a DINO~\citep{dinov2} tokenizer and alternating-attention~\citep{vggt} layers to aggregate global scene structure across views. 

\noindent\textbf{Class-agnostic Instance-grounded Geometry Transformer.}~IGGT series~\citep{iggt,iggt4d} attaches an instance head to the geometry transformer to predict instance embeddings $\{\mathbf{F}_{i}\in\mathbb{R}^{8\times H \times W}\}_{i=1}^{N}$ besides geometry.  This instance representation is learned by contrastive learning and can yield class-agnostic entity masklets by clustering. However, the clustering-based postprocessing is time-consuming and hyperparameter-sensitive. Moreover, the instance embeddings are obtained from coarse geometry-instance feature tokens constrained by the native input resolution of about 518 pixels in width, which hinders fine-grained entity perception.

In this work, we seek an end-to-end framework for effective class-agnostic entity understanding without clustering, which directly operates on high-resolution inputs $\{I^{hr}_{i}\}_{i=1}^{N}$ for fine-grained geometry-instance representation, formulated as:
\begin{equation}
    f_{\theta}\big(\{I^{hr}_{i}\}_{i=1}^{N}\big)
    =\big\{(\mathbf{g}_{i},D_{i},M_{i}^{\mathrm{ID}})\big\}_{i=1}^{N},
    \label{eq:end_to_end}
    \end{equation}
    Here, $M_{i}^{\mathrm{ID}}\in\{0,1,\ldots,O\}^{H\times W}$ is a dense object-ID map: pixels that share the same ID $j\geq 1$ belong to the same entity across views, and $0$ denotes no-object. Unprojecting $D_{i}$ and $M_{i}^{\mathrm{ID}}$ with $\mathbf{g}_{i}$ yields a global point cloud with each 3D point assigned an entity identity.

\begin{table*}[t]
    \centering
    \vspace{-4mm}
    \caption{\textbf{Class-agnostic video entity segmentation and class-aware video panoptic segmentation.} $^\dag$We run offline inference with full global attention on IGGT4D~\citep{iggt4d} for fair comparison. $^\ddag$We additionally train an LR version of Entity3R, with HR branch removed. Runtime is measured with a 100-frame sequence on a single NVIDIA H20 GPU.}
    \label{tab:entity-seg}
    \setlength{\tabcolsep}{2.2pt}
    \renewcommand{\arraystretch}{1.05}
    \resizebox{\textwidth}{!}{%
    \begin{tabular}{l|c cc cc cc|cc cc|c}
    \toprule
    \multirow{3}{*}[-1ex]{\textbf{Method}} & \multicolumn{7}{c|}{\textbf{Video Entity Seg.}} & \multicolumn{4}{c|}{\textbf{Video Panoptic Seg.}} & \multirow{3}{*}[-1ex]{Runtime} \\
    \cmidrule(lr){2-8} \cmidrule(lr){9-12}
    & \textbf{DAVIS} & \multicolumn{2}{c}{\textbf{VIPSeg}} & \multicolumn{2}{c}{\textbf{Hypersim}} & \multicolumn{2}{c|}{\textbf{Waymo}}
    & \multicolumn{2}{c}{\textbf{VIPSeg}} & \multicolumn{2}{c|}{\textbf{Waymo}} & \\
    \cmidrule(lr){2-2} \cmidrule(lr){3-4} \cmidrule(lr){5-6} \cmidrule(lr){7-8}
    \cmidrule(lr){9-10} \cmidrule(lr){11-12}
    & $\mathcal{J}\&\mathcal{F}$ $\uparrow$
    & VEQ $\uparrow$ & $\mathrm{STQ}_{\mathrm{EN}}$ $\uparrow$
    & VEQ $\uparrow$ & $\mathrm{STQ}_{\mathrm{EN}}$ $\uparrow$
    & VEQ $\uparrow$ & $\mathrm{STQ}_{\mathrm{EN}}$ $\uparrow$
    & VPQ $\uparrow$ & STQ $\uparrow$
    & VPQ $\uparrow$ & STQ $\uparrow$
    & \\
    \midrule
    EntitySAM~\citep{entitysam} & \U{71.4} & 61.4 & 43.3 & 10.6 & 11.0 & 24.1 & 11.1 & \U{29.3} & 33.2 & 13.8 & 6.9 & 86s \\
    SegVGGT~\citep{segvggt} & 31.4 & 26.3 & 29.4 & 11.8 & 16.1 & 8.3 & 11.2 & 13.9 & 20.5 & 3.8 & 6.2 & 21s \\
    IGGT~\citep{iggt} & 37.0 & 44.5 & 41.2 & 11.2 & 17.4 & 11.9 & 10.5 & 21.3 & 28.7 & 8.2 & 5.2 & OOM \\
    IGGT4D~\citep{iggt4d}$^\dag$ & 59.9 & 58.1 & 48.1 & \U{24.7} & \U{28.0} & 27.1 & 17.5 & 27.8 & 35.2 & 15.2 & 10.7 & 36s \\
    \textbf{Ours} (w/o HR branch)$^\ddag$ & 66.5 & \U{63.7} & \U{49.5} & 22.8 & 22.7 & \U{27.9} & \U{17.7} & 28.7 & \U{36.4} & \U{16.0} & \U{11.9} & \B{9s} \\
    \textbf{Ours} & \B{78.9} & \B{71.2} & \B{53.6} & \B{34.3} & \B{31.6} & \B{40.3} & \B{22.9} & \B{31.7} & \B{40.0} & \B{20.8} & \B{16.0} & \U{19s} \\
    \bottomrule
    \end{tabular}}
\end{table*}

\begin{figure}[t]
\centering
\includegraphics[width=\linewidth]{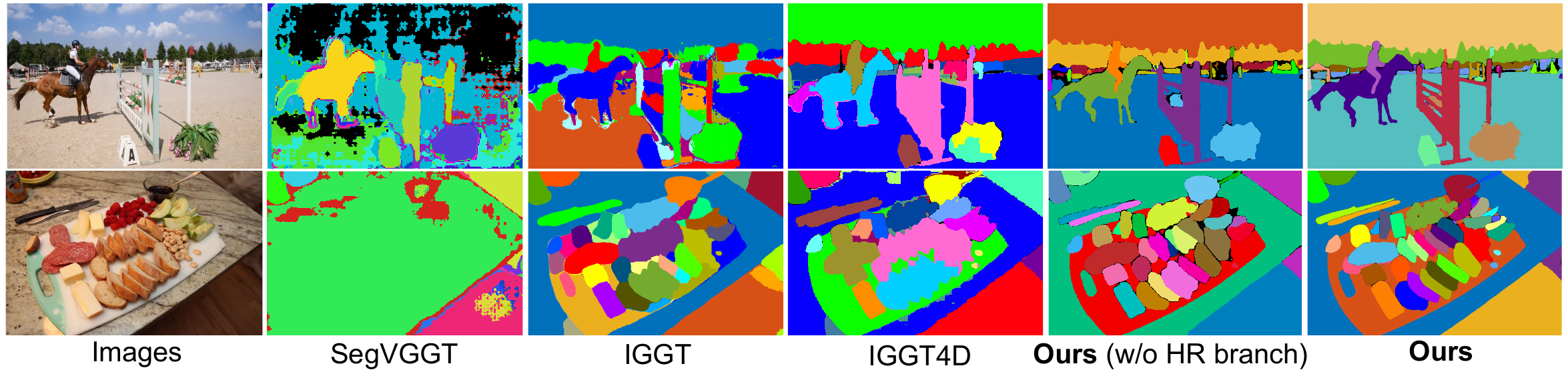}
\vspace{-4mm}
\caption{\textbf{Qualitative comparison of entity segmentation.}~Entity3R produces higher-quality entity masks than others, and our high-resolution (HR) design yields more accurate object boundaries.}
\vspace{-2mm}
\label{fig:teaser2}
\end{figure}

\subsection{Fine-grained Geometry-Instance Representation}
\label{sec:backbone}

Directly processing high-resolution images with a geometry transformer is prohibitively expensive, as global attention constitutes the computational bottleneck. Inspired by DAGE~\citep{dage}, we use a dual-branch backbone with high and low resolutions. The low-resolution (LR) branch follows the same procedure as geometry transformer to extract LR feature tokens $\{T^{lr}_{i}\}_{i=1}^{N}$ and camera tokens $\{T^{cam}_{i}\}_{i=1}^{N}$ from downsampled LR frames $\{I^{lr}_{i}\}_{i=1}^{N}$. The camera tokens are then fed to a camera head to predict camera parameters $\{\mathbf{g}_{i}\}_{i=1}^{N}$. 

The High-resolution (HR) branch is responsible for extracting fine-grained semantic features and preserving object boundaries by encoding each HR frame independently. Since the DINO series~\citep{dinov2,dinov3} provides strong dense representations for downstream depth estimation and semantic segmentation tasks, we directly leverage its pretrained priors and adopt it as the HR backbone to extract per-frame HR feature tokens $\{T^{hr}_{i}\}_{i=1}^{N}$. In this way, we keep the HR tokens out of the global attention layers, which reduces the computational cost significantly. 

To fuse the HR tokens with the LR tokens such that the final representation is both fine-grained and globally consistent, we utilize a \textit{frame-wise adaptor} in which the HR tokens serve as queries and the LR tokens as keys/values, allowing the HR tokens to retain fine-grained semantics while absorbing global scene information from the LR tokens. The adaptor is a lightweight network following DAGE with 3 transformer blocks, each structured as a cross-attention layer ($\mathrm{CAttn}$) followed by a self-attention layer ($\mathrm{SAttn}$) and a feed-forward MLP network ($\mathrm{FFN}$), and formulated as:
\begin{equation}
T^{fuse}_{i}=\mathrm{FFN}(\mathrm{SAttn}(\mathrm{CAttn}(Q=T^{hr}_{i},K=T^{lr}_{i},V=T^{lr}_{i}))),
\label{eq:adaptor}
\end{equation}
where $\{T^{fuse}_{i}\}_{i=1}^{N}$ serve as the \textit{fine-grained geometry-instance representation} for the subsequent dense prediction heads. The depth head can predict HR depth maps $\{D_{i}\}_{i=1}^{N}$ from $\{T^{fuse}_{i}\}_{i=1}^{N}$ following the standard design in geometry transformers.

\subsection{End-to-end Entity Prediction}
\label{sec:prediction}

\begin{table*}[t]
    \centering
    \vspace{-4mm}
    \caption{\textbf{Camera pose estimation.}~We report pose AUC under angular error thresholds of 5/15 degrees for RealEstate10K~\citep{realestate10k} and Co3Dv2~\citep{co3dv2}, and distance errors of rotation/translation on Sintel~\citep{sintel}, TUM-dynamics~\citep{tum}, and ScanNet~\citep{scannet}. All methods have seen Co3Dv2 and ScanNet in training.
}
    \label{tab:pose}
    \setlength{\tabcolsep}{2.0pt}
    \renewcommand{\arraystretch}{1.05}
    \resizebox{\textwidth}{!}{%
    \begin{tabular}{l|cc cc|ccc ccc ccc}
    \toprule
    \multirow{2}{*}[-1ex]{\textbf{Method}} & \multicolumn{2}{c}{\textbf{RealEstate10K}} & \multicolumn{2}{c|}{\textbf{Co3Dv2}}
    & \multicolumn{3}{c}{\textbf{Sintel}} & \multicolumn{3}{c}{\textbf{TUM-dynamics}} & \multicolumn{3}{c}{\textbf{ScanNet}} \\
    \cmidrule(lr){2-3} \cmidrule(lr){4-5} \cmidrule(lr){6-8} \cmidrule(lr){9-11} \cmidrule(lr){12-14}
    & AUC/$5$ $\uparrow$ & AUC/$15$ $\uparrow$
    & AUC/$5$ $\uparrow$ & AUC/$15$ $\uparrow$
    & ATE $\downarrow$ & RPE-t $\downarrow$ & RPE-r $\downarrow$
    & ATE $\downarrow$ & RPE-t $\downarrow$ & RPE-r $\downarrow$
    & ATE $\downarrow$ & RPE-t $\downarrow$ & RPE-r $\downarrow$ \\
    \midrule
    DA3~\citep{da3} & 61.02 & 79.47 & \B{72.81} & \B{86.70} & 0.117 & 0.043 & 0.336 & 0.012 & \U{0.009} & 0.312 & 0.033 & \B{0.013} & 0.636 \\
    $\pi^{3}$~\citep{pi3} & 63.03 & 80.10 & 68.46 & 84.27 & 0.073 & 0.038 & 0.288 & 0.014 & \U{0.009} & 0.307 & \B{0.030} & \B{0.013} & \B{0.345} \\
    VGGT~\citep{vggt} & 30.71 & 60.01 & 67.83 & 83.95 & 0.171 & 0.062 & 0.471 & 0.012 & 0.010 & 0.309 & 0.035 & 0.015 & \U{0.376} \\
    VGGT-$\Omega$~\citep{vggtomega} & 58.28 & 78.54 & 60.12 & 80.44 & \U{0.040} & \U{0.027} & \B{0.239} & \B{0.007} & \B{0.006} & \B{0.279} & 0.033 & \U{0.014} & 0.386 \\
    DAGE~\citep{dage} & 52.49 & 74.67 & 62.90 & 81.62 & 0.141 & 0.054 & 0.416 & 0.014 & 0.011 & 0.323 & \U{0.031} & \B{0.013} & 0.386 \\
    SegVGGT~\citep{segvggt} & 19.80 & 50.78 & 57.64 & 79.40 & 0.109 & 0.053 & 0.442 & 0.012 & \U{0.009} & 0.314 & 0.035 & 0.015 & 0.382 \\
    IGGT~\citep{iggt} & 41.11 & 68.26 & 67.71 & 83.94 & 0.167 & 0.058 & 0.469 & 0.013 & 0.011 & 0.313 & 0.034 & 0.015 & 0.379 \\
    IGGT4D~\citep{iggt4d} & \U{64.37} & \U{80.95} & \U{70.14} & \U{85.31} & 0.096 & 0.045 & 0.328 & 0.014 & 0.011 & 0.321 & 0.032 & \B{0.013} & 0.387 \\
    \textbf{Ours} & \B{65.27} & \B{81.64} & 66.18 & 83.30 & \B{0.038} & \B{0.024} & \U{0.249} & \U{0.008} & \B{0.006} & \U{0.282} & 0.033 & 0.015 & 0.388 \\
    \bottomrule
    \end{tabular}}
    \vspace{-4mm}
\end{table*}

\noindent\textbf{End-to-end Entity Prediction.}~We introduce $O$ learnable object queries $\{\mathbf{o}_{j}\in\mathbb{R}^{C}\}_{j=1}^{O}$ to detect entities in the scene, avoiding the hand-crafted and time-consuming clustering used by prior instance-grounded models~\citep{iggt, iggt4d}. To this end, we use Mask2Former-style~\citep{mask2former} pixel decoder, based on multi-scale deformable attention~\citep{deformabledetr}, as our instance head to map the fused fine-grained tokens $\{T^{fuse}_{i}\}_{i=1}^{N}$ into multi-scale instance-specific features $\{F_{i}^{s}\}_{i=1}^{N}$, where $s\in\{32,16,8,4\}$ and $F_{i}^{s}\in\mathbb{R}^{C\times\frac{H}{s}\times\frac{W}{s}}$. The instance features inherit the global consistency from the fine-grained geometry-instance representation. We therefore concatenate them across frames into a global scene context $F^{s}=\mathrm{Concat}(F_{1}^{s},\ldots,F_{N}^{s})\in\mathbb{R}^{C\times (N\times\frac{H}{s}\times\frac{W}{s})}$ for \textit{full-context querying}. Following Mask2Former, our entity decoder updates the object queries using six transformer blocks that alternately attend to the intermediate features $F^{32}$, $F^{16}$, and $F^{8}$ over two rounds. Each block is formulated as:
\begin{equation}
\mathbf{o}_{j}=\mathrm{FFN}(\mathrm{SAttn}(\mathrm{CAttn}(Q=\mathbf{o}_{j},K=F^{s},V=F^{s}))).
\label{eq:object_query}
\end{equation}
where $s$ takes values in $\{32,16,8,32,16,8\}$ in order. After each block, the updated queries (including the initial queries) are dotted with $F^{4}$ to produce entity mask probabilities, $\mathrm{sigmoid}(\mathbf{o}_{j}^{\top} F^{4})$, which provide deep supervision and are binarized into hard attention masks for the next block. Each query is also fed to an MLP binary classifier that predicts an entityness value $p_{j}\in\{0,1\}$, indicating whether it corresponds to an object or to no-object. The dot product of the fully updated queries with $F^{4}$ yields the final results, which are split, reshaped, and upsampled into per-frame HR predictions and then resolved into dense object-ID maps $\{M_{i}^{\mathrm{ID}}\}_{i=1}^{N}$ under a non-overlapping constraint.

\noindent\textbf{Chunk-wise Object Decoding to Boost Inference Scalability.}~Above full-context querying procedure lets every object query attend to the full scene context $F^{s}$ over all $N$ frames at once. This raises two issues. First, decoding over $F^{s}$ makes memory grow with $N$. Second, as more frames bring more entities, a fixed query set gradually fails to cover them all. We therefore propose a memory-efficient \textit{chunk-wise object decoding} strategy that is an optional plug-in to boost \textit{inference scalability} without modifying or retraining the model, as illustrated in the bottom row of Fig.~\ref{fig:overview}. Specifically, we partition the $N$ frames into several non-overlapping chunks and predict entities chunk by chunk. Within each chunk, the queries attend only to that chunk's concatenated features. This bounds both the attended context memory and the number of entities per pass, while still producing cross-frame-consistent masklets inside each chunk. To keep entity IDs across chunks, we perform inter-chunk association by pooling each masklet over $F^{4}$ into a normalized feature center, and then matching centers in the current chunk to all previously established entity centers via cosine similarity. Matched centers are assigned to the corresponding entity IDs, while unmatched ones are treated as new entities and assigned new IDs. Note that the chunk-wise method applies only to object decoding, and all frames still enter the backbone (LR geometry transformer) jointly, which preserves the cross-frame consistency of the instance features $F^{4}$ and thereby ensures accurate inter-chunk association.

\subsection{Model Training}
\label{sec:obj}

\begin{figure*}[t]
    \centering
    \vspace{-2mm}
    \includegraphics[width=\textwidth]{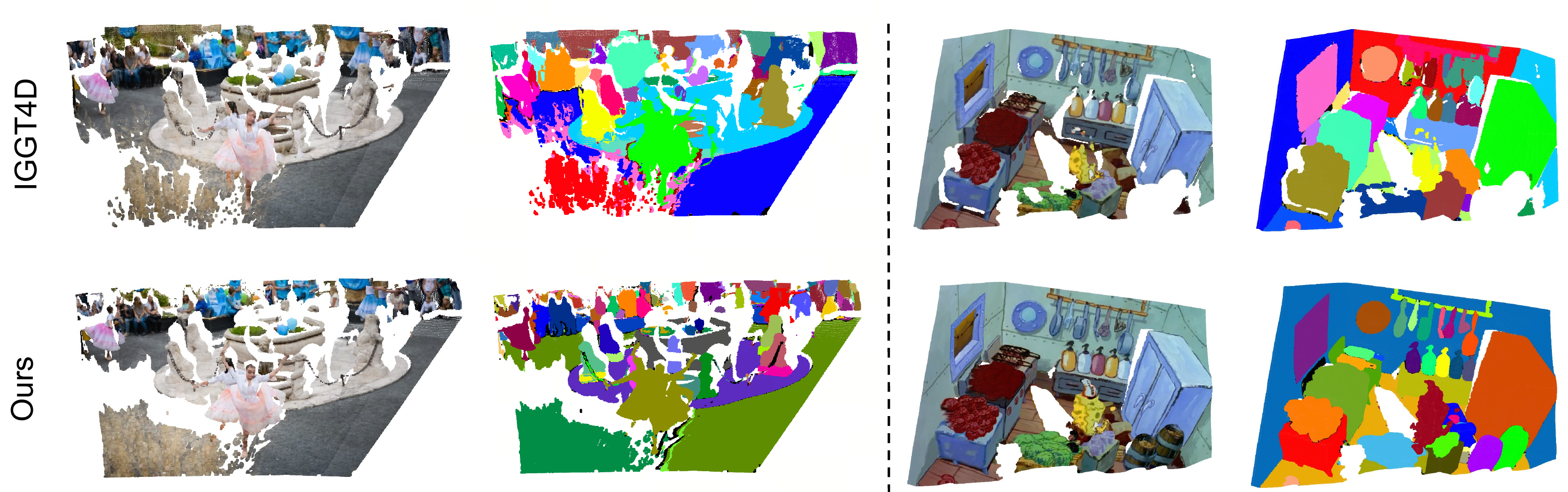}
    \vspace{-6mm}
    \caption{\textbf{Qualitative comparison of entity-colored point clouds.}
    For each scene we show the reconstructed point cloud and the corresponding entity identities.
    Compared with IGGT4D~\citep{iggt4d}, Entity3R produces denser geometry with fewer holes and more accurate entity regions.}
    \label{fig:pc_vis}
\end{figure*}

We train Entity3R with the combination of geometry and entity supervisions: $\mathcal{L}=\mathcal{L}_{\mathrm{geo}}+\mathcal{L}_{\mathrm{ent}}$. 

\noindent\textbf{Geometry Loss.} $\mathcal{L}_{\mathrm{geo}}$ follows the standard design in geometry transformers (see in Sec.~\ref{sec:suppl_model_training}).

\noindent\textbf{Entity Loss.}~The entity loss comprises three terms: $\mathcal{L}_{\mathrm{ent}}=\lambda_{\mathrm{bce}}\mathcal{L}_{\mathrm{bce}}+\lambda_{\mathrm{dice}}\mathcal{L}_{\mathrm{dice}}+\lambda_{\mathrm{cls}}\mathcal{L}_{\mathrm{cls}}$. Following query-based image detection/segmentation~\citep{detr,maskformer,mask2former}, we first match predicted masklets to ground-truth entity masklets with the Hungarian algorithm~\citep{hungarian}. For matched pairs, $\mathcal{L}_{\mathrm{bce}}$ and $\mathcal{L}_{\mathrm{dice}}$ supervise the masklets with binary cross-entropy and Dice losses. $\mathcal{L}_{\mathrm{cls}}$ is a binary classification loss on entityness as in~\citep{entityseg,cropformer}, where matched queries are trained as objects and unmatched queries as no-object. In this way, supervision is class-agnostic and uses no category vocabulary. We apply the entity loss with deep supervision~\citep{mask2former} to the object queries after they are updated by each block in entity decoder.

\noindent\textbf{Implementation Details.}~Entity3R processes HR inputs at a width of 1024 and downsamples them to a width of 512 for LR branch. We initialize the LR geometry transformer, camera head and depth head from VGGT-$\Omega$~\citep{vggtomega}, as its scalable training yields a stronger geometric backbone and its register attention improves training and inference efficiency. The HR branch uses a DINOv3 ViT-L/16~\citep{dinov3} encoder, followed by a 3-block frame-wise adaptor. The instance head is a Mask2Former-style pixel decoder based on multi-scale deformable attention~\citep{deformabledetr}, with a {ViT}-Adapter~\citep{vitadapter} to extract multi-scale features. We prepare 200 learnable object queries to be updated by the 6-block entity decoder with masked attention. More details are provided in Sec.~\ref{sec:suppl_model_training}.

\section{EntityWorld Dataset}

\noindent\textbf{EntityWorld.} Joint geometry reconstruction and entity perception need both geometric supervision and 3D/4D-consistent entity masklets at scale. We therefore contribute \textbf{EntityWorld}, a new dataset that contains nearly 84.8M frames and 360K sequences with cameras, depths, and class-agnostic entity masklets. EntityWorld is constructed by combining existing geometric datasets and our newly collected photorealistic game data. We follow OmniWorld~\citep{omniworld} to collect game sequences. Depth is accessed during rendering with ReShade~\citep{reshade}, and synchronized RGB frames are captured from the screen with OBS~\citep{OBS}, yielding about 14.3M frames and 143K game sequences with accurate geometry labels. Game data can span diverse scene types including indoor, outdoor-urban, and outdoor-natural scenes, and thus complement the community sources that are dominated by indoor and outdoor-urban scenes. The full dataset composition of EntityWorld is summarized in Sec.~\ref{sec:suppl_entityworld_composition} and Tab.~\ref{tab:entityworld_sources}.

\noindent\textbf{Automatic Entity Mask Annotation.} Some of existing synthetic datasets already provide the entity mask GTs by the rendering engines and thus can be directly used. For the rest, we propose an automatic annotation pipeline for scalable entity masklet generation. Specifically, given RGB frames, we initialize entity mask proposals using CropFormer~\citep{cropformer} on the first frame. Compared to auto-segmentation in SAM series~\citep{sam, sam2} used by IGGT~\citep{iggt} labeling, CropFormer can produce higher-quality entity masks on an image without over-/under-segmentation (see the label comparison in Fig.~\ref{fig:suppl_label_cmp} and Tab.~\ref{tab:annotation_assessment}). We then propagate the proposals temporally with SAM2~\citep{sam2}. At each frame, we re-detect entities in the uncovered regions with CropFormer, while using an optical flow model~\citep{raft} to determine whether a re-detected entity corresponds to a lost SAM2 track or a newly appearing object. We also use the optical flow model to detect large propagation jumps between adjacent frames, which may indicate an abrupt scene change. In such cases, we split the sequence at the jump and start a new clip from this frame, repeating the above procedure. We finally drop the sequences with less than 24 frames. The annotation pipeline and the dataset statistics are presented in Fig.~\ref{fig:entityworld}. 
\section{Experiments}

\subsection{Video Entity Segmentation}
We evaluate Entity3R's basic entity perception ability on the video entity segmentation task, reporting the metrics VEQ and $\mathrm{STQ}^{\mathrm{EN}}$ proposed in EntitySAM~\citep{entitysam}, which are class-agnostic variants of the VPQ~\citep{kim2020vps} and STQ~\citep{weber2021step} metrics for video panoptic segmentation. Considering that the ground-truth annotations are not exhaustive for all objects in the scene, we do not penalize false positive tracks directly in the metrics, as in~\cite{deva}. However, false positive tracks are indirectly penalized by requiring the output mask to be mutually exclusive. We adopt DAVIS~\citep{davis2017}, VIPSeg~\citep{vipseg}, Hypersim~\citep{hypersim}, and Waymo~\citep{waymo} for evaluation. Since DAVIS is a video object segmentation dataset, we report its widely used $\mathcal{J}\&\mathcal{F}$~\citep{davis2016} metrics instead. The comparison baseline models include EntitySAM, SegVGGT~\citep{segvggt}, IGGT~\citep{iggt}, and IGGT4D~\citep{iggt4d}. The first is a video entity segmentation model based on SAM2~\citep{sam2} while the remaining are instance-grounded geometry transformers. To establish fair comparison, we additionally train an LR version of Entity3R, with HR branch removed. Results are shown in Tab.~\ref{tab:entity-seg}. Our LR model already surpasses the baselines on most metrics with the lowest runtime, while the stronger HR model comprehensively outperforms all methods at a still highly competitive inference speed. We present qualitative entity comparisons in Fig.~\ref{fig:teaser2} and Fig.~\ref{fig:pc_vis}, demonstrating Entity3R's superior entity masklet quality and object boundary accuracy. More qualitative results are available in the Sec.~\ref{sec:suppl_qualitative}.

\subsection{Downstream Video Panoptic Segmentation}
Class-agnostic entity segmentation can be further applied to downstream open-vocabulary semantic understanding. Following IGGT~\citep{iggt}, we use the predicted entity masklets as prompts to an off-the-shelf VLM~\citep{lseg}. Vision-language features are average-pooled within each masklet and matched against text embeddings to assign a semantic category, yielding class-aware video panoptic predictions. We then evaluate on this video panoptic segmentation task, reporting VPQ/STQ metrics on VIPSeg and Waymo that contain category labels. As shown in Tab.~\ref{tab:entity-seg}, our method again leads on both datasets, indicating that our stronger entity perception also benefits class-aware semantic understanding.

\subsection{Geometry Estimation}
We assess Entity3R's geometry reconstruction ability following the same evaluation protocol of $\pi^{3}$~\citep{pi3} and ZipMap~\citep{zipmap}, covering camera pose estimation, video depth estimation, and point map estimation. For camera pose, we evaluate on RealEstate10K~\citep{realestate10k}, Co3Dv2~\citep{co3dv2}, Sintel~\citep{sintel}, TUM-dynamics~\citep{tum}, and ScanNet~\citep{scannet} in Tab.~\ref{tab:pose}. For video depth, we evaluate on Sintel~\citep{sintel}, KITTI~\citep{kitti}, and Bonn~\citep{bonn} in Tab.~\ref{tab:videodepth}. For point map estimation, we evaluate on 7-Scenes~\citep{sevenscenes}, NRGBD~\citep{nrgbd}, ETH3D~\citep{eth3d}, and DTU~\citep{dtu} in Tabs.~\ref{tab:pointmap-eth-dtu}--\ref{tab:pointmap}. We can see that Entity3R achieves performance comparable to state-of-the-art geometry models on most metrics across these benchmarks, while jointly providing entity perception.

\begin{table*}[t]
\centering
\vspace{-1mm}
\caption{\textbf{Ablation study results on object query placement and chunk-wise decoding.} $S$ denotes the chunk size. Max \#Frames is maximum number of frames that can be processed in an NVIDIA H20-96GB GPU. Runtime is measured with a 400-frame sequence on H20.}
\vspace{-2mm}
\label{tab:ablation}
\small
\setlength{\tabcolsep}{2.8pt}
\renewcommand{\arraystretch}{1.05}
\resizebox{\textwidth}{!}{%
\begin{tabular}{c l | cccc | cccc | c c}
\toprule
& \multirow{2}{*}[-0.4em]{\textbf{Method}}
& \multicolumn{4}{c|}{\textbf{VIPSeg}~(Shorter Seqs.)}
& \multicolumn{4}{c|}{\textbf{Waymo}~(Longer Seqs.)}
& \multirow{2}{*}[-0.4em]{\shortstack{Max\\\#Frames}}
& \multirow{2}{*}[-0.4em]{\shortstack{Runtime\\@400F}} \\
\cmidrule(lr){3-6} \cmidrule(lr){7-10}
&
& VEQ $\uparrow$ & $\mathrm{STQ}_{\mathrm{EN}}$ $\uparrow$ & VPQ $\uparrow$ & STQ $\uparrow$
& VEQ $\uparrow$ & $\mathrm{STQ}_{\mathrm{EN}}$ $\uparrow$ & VPQ $\uparrow$ & STQ $\uparrow$
& & \\
\midrule
(a) & Inside-aggregation query interaction
& 64.8 & 48.2 & 27.5 & 35.1 & 33.6 & 18.4 & 16.2 & 11.8 & -- & -- \\
(b) & \textbf{Post-aggregation query interaction (Ours)}
& \B{71.2} & \B{53.6} & \B{31.7} & \B{40.0} & \B{40.3} & \B{22.9} & \B{20.8} & \B{16.0} & -- & -- \\
\midrule
(c) &Chunk-wise object decoding ($S{=}1$)
& \B{71.4} & \B{53.6} & 31.7 & \B{40.1} & \B{42.1} & \B{23.5} & \B{21.2} & \B{16.5} & \B{1200} & 155s \\
(d) &Chunk-wise object decoding ($S{=}2$)
& \B{71.4} & \B{53.6} & 31.7 & \B{40.1} & 41.9 & \B{23.5} & 20.9 & 16.4 & \B{1200} & 147s \\
(e) &Chunk-wise object decoding ($S{=}4$)
& \B{71.4} & 53.5 & 31.7 & \B{40.1} & 41.7 & 23.4 & 21.0 & \B{16.5} & \B{1200} & 143s \\
(f) &Chunk-wise object decoding ($S{=}8$)
& 71.3 & 53.5 & \B{31.8} & 40.0 & 41.5 & 23.3 & 20.9 & 16.3 & \B{1200} & 141s \\
(g) &Chunk-wise object decoding ($S{=}16$)
& 71.3 & 53.5 & 31.7 & 40.0 & 41.2 & 23.1 & 20.7 & 16.3 & \B{1200} & 141s \\
(h) & Full-context object decoding (w/o chunking)
& 71.2 & \B{53.6} & 31.7 & 40.0 & 40.3 & 22.9 & 20.8 & 16.0 & 400 & \textbf{140s} \\
\bottomrule
\end{tabular}}
\end{table*}

\subsection{Ablation Studies}
\noindent
\begin{minipage}[t]{0.45\textwidth}
\textbf{Effect of Object Query Placement.}~We study two designs for how object queries interact with instance features: \textit{inside-aggregation}, where object queries interact within the aggregation layers of the geometry transformer, following SegVGGT~\citep{segvggt}, an indoor segmentation method, and \textit{post-aggregation}, our design where object queries are updated in a separate entity decoder after aggregation. We ablate this design choice in Tab.~\ref{tab:ablation}(a,b), showing that post-aggregation querying is clearly superior. Features inside the aggregation layers are still primarily geometric and not yet instance-specific, so early query interaction struggles to form reliable entity representations. In contrast, our entity decoder updates the queries with instance-specific features produced by the instance head, enabling more effective entity decoding.
\par\smallskip
\end{minipage}%
\hfill
\begin{minipage}[t]{0.52\textwidth}
\centering
\vspace{-4mm}
\includegraphics[width=\linewidth]{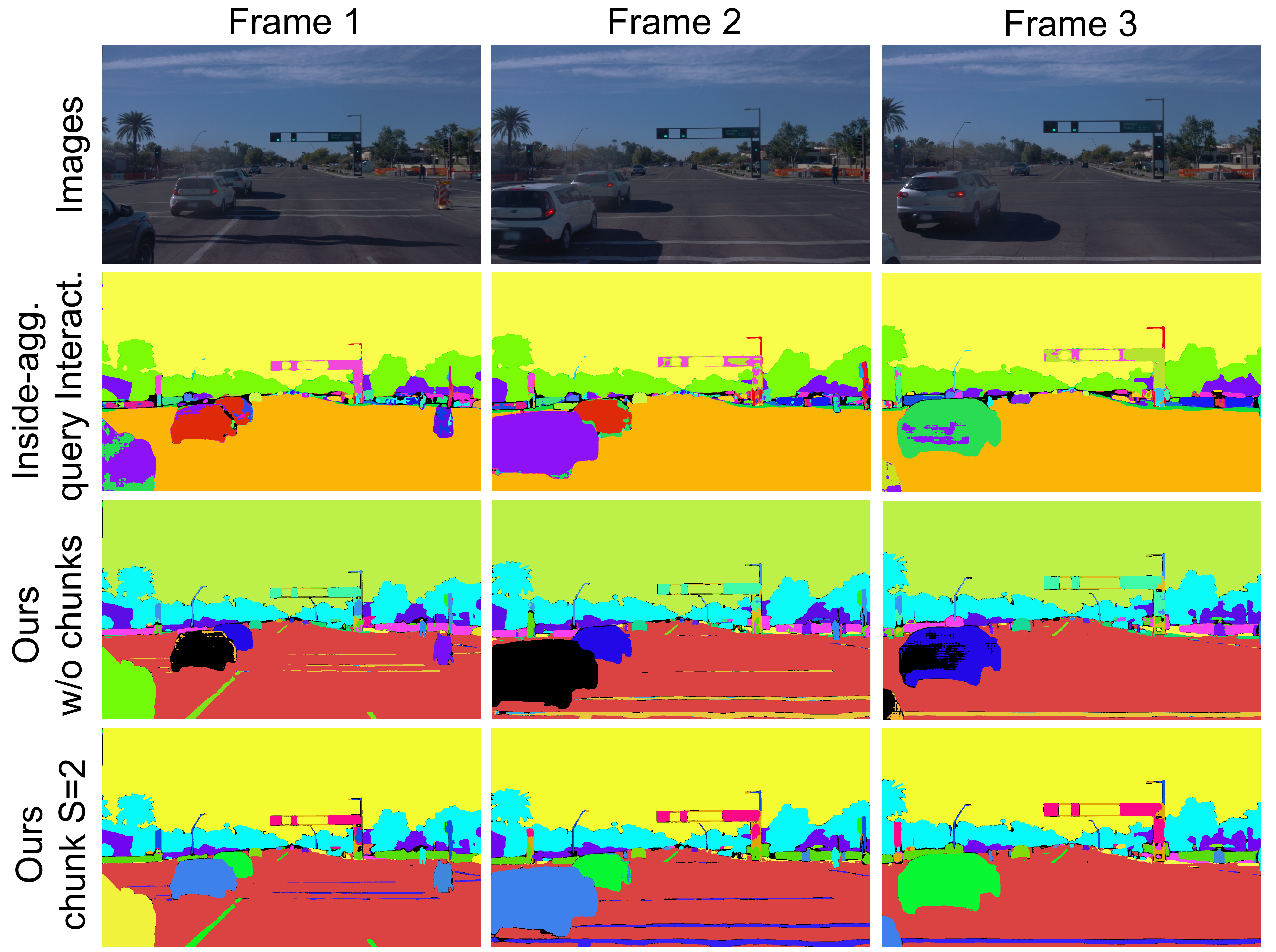}
\vspace{-2mm}
\captionsetup{hypcap=false}
\captionof{figure}{\textbf{Qualitative ablation results.}~Our chunk-wise object decoding can handle longer sequences with more entities than a fixed set of queries can represent.}
\label{fig:ablation_qual}
\end{minipage}

\noindent\textbf{Effect of Chunk-wise Object Decoding.}~We explore the influence of chunk-wise object decoding at inference. As listed in Tab.~\ref{tab:ablation}(c--h), a small chunk size ($S{=}1$ or $S{=}2$) works best. On shorter sequences such as VIPSeg, chunking preserves accuracy without degradation. On longer sequences such as Waymo, splitting into chunks even improves performance by covering more entities under a fixed query budget. Chunk-wise decoding also supports more frames: full-context decoding OOMs beyond 400 frames, whereas chunk-wise alternatives can process up to 1200 frames in a memory-efficient manner. Moreover, the additional runtime cost introduced by inter-chunk association remains acceptable compared with full-context decoding. We also provide the qualitative results in Fig.~\ref{fig:ablation_qual}.

\noindent\textbf{Effect of data source.}~We decompose our EntityWorld dataset into three subsets: public data with existing mask labels, public data annotated by our pipeline, and the newly collected EntityWorld-Game. As listed in Tab.~\ref{tab:data_ablation}(a), starting from public data with existing mask labels, adding public data annotated by our pipeline consistently improves the performance, and further adding EntityWorld-Game (i.e., full EntityWorld) yields the best results. This shows that both our annotated public sources and the newly collected EntityWorld-Game data contribute to the final gains.

\noindent\textbf{Effect of data scale.}~We subsample EntityWorld by taking 25\%/50\%/75\%/100\% of sequences from each source dataset at the sequence level, so that the relative composition across sources is preserved. Performance improves monotonically with the sampling ratio. As shown in Tab.~\ref{tab:data_ablation}(b), results at 75\% already approach the full-data setting, while 100\% EntityWorld remains best, indicating that EntityWorld’s scale is beneficial.

\noindent\textbf{Effect of architecture choice.}~
We perform architecture ablation (object query vs. clustering), with fixed training data and backbone. As shown in Tab.~\ref{tab:data_ablation}(c), object-query decoding consistently outperforms clustering across all reported metrics. This shows that the gains of Entity3R are not solely due to EntityWorld or backbone initialization; the object-query architecture itself contributes substantial improvements. 

\noindent\textbf{Effect of backbone choice.}~We conduct backbone ablation (VGGT vs. VGGT-$\Omega$), with fixed training data and architecture. Tab.~\ref{tab:data_ablation}(d) shows that VGGT-$\Omega$ backbone consistently outperforms VGGT across all metrics. Although backbone choice is not our direct contribution, this comparison demonstrates that joint geometry–instance learning benefits from a stronger geometry-pretrained backbone.

\noindent\textbf{Effect of freezing encoders.}~We also conduct an experiment that freezes the encoders and trains only the fusion adapter and decoder heads. As shown in Tab.~\ref{tab:data_ablation}(e), freezing the encoders leads to a clear drop across all tracks. This indicates that adapting encoder representations is necessary for learning a unified geometry–instance representation, rather than treating the pretrained encoders as fixed feature extractors and only learning the adapter/heads.

\begin{table*}[t]
    \centering
    \caption{ \textbf{Ablation on (a) training data source, (b) training data scale, (c) architecture choice, (d) backbone choice, and (e) freezing encoders.}}
    \vspace{-2mm}
    \label{tab:data_ablation}
    \small
    \setlength{\tabcolsep}{4.5pt}
    \renewcommand{\arraystretch}{1.05}
    \resizebox{\textwidth}{!}{%
    \begin{tabular}{l | cccc | cccc }
    \toprule
    \multirow{2}{*}[-0.4em]{\textbf{Method}}
    & \multicolumn{4}{c|}{\textbf{VIPSeg}~(Shorter Seqs.)}
    & \multicolumn{4}{c}{\textbf{Waymo}~(Longer Seqs.)} \\
    \cmidrule(lr){2-5} \cmidrule(lr){6-9}
    & VEQ $\uparrow$ & $\mathrm{STQ}_{\mathrm{EN}}$ $\uparrow$ & VPQ $\uparrow$ & STQ $\uparrow$
    & VEQ $\uparrow$ & $\mathrm{STQ}_{\mathrm{EN}}$ $\uparrow$ & VPQ $\uparrow$ & STQ $\uparrow$ \\
    \midrule
    \multicolumn{9}{l}{\textit{(a) Data source}} \\
    \quad Public data with existing mask labels
    & 63.8 & 47.6 & 26.4 & 34.6 & 34.8 & 18.3 & 16.8 & 12.9 \\
    \quad + Public data annotated in this work
    & 67.5 & 50.3 & 29.1 & 37.4 & 37.2 & 20.5 & 18.6 & 14.7 \\
    \quad + EntityWorld-Game (full EntityWorld)
    & \textbf{71.2} & \textbf{53.6} & \textbf{31.7} & \textbf{40.0}
    & \textbf{40.3} & \textbf{22.9} & \textbf{20.8} & \textbf{16.0} \\
    \midrule
    \multicolumn{9}{l}{\textit{(b) Data scale}} \\
    \quad 25\% EntityWorld
    & 64.2 & 48.1 & 27.0 & 34.9 & 35.3 & 18.6 & 17.0 & 13.3 \\
    \quad 50\% EntityWorld
    & 67.9 & 50.4 & 28.9 & 37.5 & 37.7 & 20.2 & 18.5 & 14.3 \\
    \quad 75\% EntityWorld
    & 70.1 & 52.7 & 30.9 & 39.1 & 39.2 & 22.1 & 20.1 & 15.4 \\
    \quad 100\% EntityWorld
    & \textbf{71.2} & \textbf{53.6} & \textbf{31.7} & \textbf{40.0}
    & \textbf{40.3} & \textbf{22.9} & \textbf{20.8} & \textbf{16.0} \\
    \midrule
    \multicolumn{9}{l}{\textit{(c) Architecture choice}} \\
    \quad Ours (clustering)
    & 62.6 & 51.1 & 30.0 & 38.7 & 31.1 & 19.8 & 18.0 & 13.5 \\
    \quad Ours (object query)
    & \textbf{71.2} & \textbf{53.6} & \textbf{31.7} & \textbf{40.0}
    & \textbf{40.3} & \textbf{22.9} & \textbf{20.8} & \textbf{16.0} \\
    \midrule
    \multicolumn{9}{l}{\textit{(d) Backbone choice}} \\
    \quad Ours (VGGT)
    & 69.7 & 52.1 & 30.6 & 39.1 & 37.9 & 21.3 & 19.2 & 14.8 \\
    \quad Ours (VGGT-$\Omega$)
    & \textbf{71.2} & \textbf{53.6} & \textbf{31.7} & \textbf{40.0}
    & \textbf{40.3} & \textbf{22.9} & \textbf{20.8} & \textbf{16.0} \\
    \midrule
    \multicolumn{9}{l}{\textit{(e) Freezing encoders}} \\
    \quad Ours (freezing encoders)
    & 63.3 & 49.7 & 28.2 & 34.7 & 35.1 & 18.2 & 17.3 & 13.4 \\
    \quad Ours
    & \textbf{71.2} & \textbf{53.6} & \textbf{31.7} & \textbf{40.0}
    & \textbf{40.3} & \textbf{22.9} & \textbf{20.8} & \textbf{16.0} \\
    \bottomrule
    \end{tabular}}
    \vspace{-3mm}
    \end{table*}

\section{Conclusion}
We presented \textbf{Entity3R}, an end-to-end entity-aware geometric foundation model that jointly recovers class-agnostic entity identities with underlying scene geometry from multi-view images or videos. Entity3R learns fine-grained geometry-instance representations with a dual-branch backbone and decodes 3D/4D-consistent entity masks via learnable object queries, without clustering. A training-free chunk-wise strategy further improves scalability to support longer sequences. We further construct \textbf{EntityWorld}, a large-scale corpus of nearly 84.8M frames and 360K sequences for joint geometry and instance learning. Experiments show that Entity3R achieves state-of-the-art entity segmentation while remaining competitive on geometry reconstruction.

\section*{AI use statement}
In this work, we used generative AI tools to implement methods, assist with translation, clean and reformat dataset, support qualitative and thematic data analysis, and interpret results.
We have not used generative AI tools to generate synthetic datasets, or to design or provide feedback on research methodology or experiments.
Helping develop theoretical models or conceptual frameworks, formulating mathematical claims, providing critical ingredients for proving mathematical claims, assisting in the writing of proofs, and proposing or refining hypotheses are not applicable to this work.
Additionally, we used generative AI tools for creating or modifying scientific figures or images, drafting parts of a research paper, summarizing or analysing existing literature, sourcing/searching for information, editing a research paper to improve readability, identifying relevant literature, formatting references, suggesting a structure for a research paper, and proposing a title or keywords for a research paper.
We have carefully reviewed all AI-assisted outputs: AI-generated code for method implementations was verified and tested for correctness by the authors, translations were checked against the intended meaning, cleaned and reformatted data were inspected for correctness, qualitative analyses and result interpretations were validated, figures and images were inspected for correctness, drafted and edited manuscript text was revised as needed, literature summaries and identified references were cross-checked against the original sources, and suggested structure, title and keywords were finalized by the authors.
We take responsibility for the final content of this work, including text, claims, and artifacts produced with the aid of generative AI.

\section*{Ethics statement}
This research methodology raises no ethical issues. No human subjects were involved. 

\section*{Reproducibility statement}
Our Entity3R is built by integrating the publicly available codebases of VGGT-$\Omega$~\citep{vggtomega}, pi3~\citep{pi3}, DINOv3~\citep{dinov3}, DAGE~\citep{dage}, Mask2Former~\citep{mask2former}, CropFormer~\citep{cropformer}, SAM2~\citep{sam2}, and RAFT~\citep{raft}. 
EntityWorld is constructed from publicly available geometric datasets together with our newly collected game sequences, for which we follow OmniWorld~\citep{omniworld} and use ReShade~\citep{reshade} and OBS~\citep{OBS} for depth access and RGB capture.
We include comprehensive data-preprocessing, data-annotation,training, inference, and implementation details in this paper and the appendix, which greatly facilitate reproducing our work.
We will release our codes, models, and datasets if this work is accepted.

\bibliography{iclr2027_conference}
\bibliographystyle{iclr2027_conference}

\appendix
\section{Appendix}
\label{sec:appendix}

\begin{table*}[t]
    \centering
    \caption{\textbf{Video depth estimation.}~We report results under scale-only alignment.}
    \label{tab:videodepth}
    \setlength{\tabcolsep}{2.2pt}
    \renewcommand{\arraystretch}{1.05}
    \resizebox{0.8\textwidth}{!}{%
    \begin{tabular}{l|cc cc cc}
    \toprule
    & \multicolumn{2}{c}{\textbf{Sintel}} & \multicolumn{2}{c}{\textbf{KITTI}} & \multicolumn{2}{c}{\textbf{Bonn}} \\
    \cmidrule(lr){2-3} \cmidrule(lr){4-5} \cmidrule(lr){6-7}
    \multirow{2}{*}[1.2em]{\textbf{Method}}
    & Abs Rel $\downarrow$ & $\delta_{1}$ $\uparrow$
    & Abs Rel $\downarrow$ & $\delta_{1}$ $\uparrow$
    & Abs Rel $\downarrow$ & $\delta_{1}$ $\uparrow$ \\
    \midrule
    DA3~\citep{da3} & 0.302 & 0.687 & 0.058 & 0.978 & 0.052 & 0.972 \\
    $\pi^{3}$~\citep{pi3} & 0.229 & 0.672 & \B{0.038} & \B{0.986} & 0.051 & 0.975 \\
    VGGT~\citep{vggt} & 0.295 & 0.649 & 0.073 & 0.963 & 0.055 & 0.971 \\
    VGGT-$\Omega$~\citep{vggtomega} & \B{0.101} & \B{0.911} & 0.056 & 0.979 & \U{0.044} & 0.972 \\
    DAGE~\citep{dage} & 0.274 & 0.665 & \U{0.042} & \U{0.983} & 0.055 & \U{0.978} \\
    SegVGGT~\citep{segvggt} & 0.288 & 0.674 & 0.116 & 0.907 & 0.048 & 0.977 \\
    IGGT~\citep{iggt} & 0.272 & 0.663 & 0.066 & 0.971 & 0.062 & 0.970 \\
    IGGT4D~\citep{iggt4d} & 0.212 & 0.723 & 0.056 & 0.968 & 0.053 & 0.972 \\
    \textbf{Ours} & \U{0.109} & \U{0.860} & 0.047 & 0.980 & \B{0.042} & \B{0.980} \\
    \bottomrule
    \end{tabular}}
\end{table*}

\begin{table*}[t]
\centering
\caption{\textbf{Point map estimation on ETH3D~\citep{eth3d} and DTU~\citep{dtu}.} Following $\pi^{3}$~\citep{pi3} setting, we report the mean and median of accuracy (Acc.), completeness (Comp.), and normal-consistency (NC.) for keyframes selected every 5 images.}
\label{tab:pointmap-eth-dtu}
\setlength{\tabcolsep}{2.4pt}
\renewcommand{\arraystretch}{1.05}
\resizebox{\textwidth}{!}{%
\begin{tabular}{l|cc cc cc|cc cc cc}
\toprule
& \multicolumn{6}{c|}{\textbf{ETH3D}} & \multicolumn{6}{c}{\textbf{DTU}} \\
\cmidrule(lr){2-7} \cmidrule(lr){8-13}
\textbf{Method}
& \multicolumn{2}{c}{Acc. $\downarrow$} & \multicolumn{2}{c}{Comp. $\downarrow$} & \multicolumn{2}{c|}{NC. $\uparrow$}
& \multicolumn{2}{c}{Acc. $\downarrow$} & \multicolumn{2}{c}{Comp. $\downarrow$} & \multicolumn{2}{c}{NC. $\uparrow$} \\
\cmidrule(lr){2-3} \cmidrule(lr){4-5} \cmidrule(lr){6-7}
\cmidrule(lr){8-9} \cmidrule(lr){10-11} \cmidrule(lr){12-13}
& Mean & Med. & Mean & Med. & Mean & Med.
& Mean & Med. & Mean & Med. & Mean & Med. \\
\midrule
DA3~\citep{da3} & 0.267 & 0.204 & 0.327 & 0.213 & 0.839 & 0.919 & 1.409 & 0.716 & 1.581 & 0.687 & 0.676 & 0.764 \\
$\pi^{3}$~\citep{pi3} & 0.187 & 0.125 & 0.210 & 0.129 & 0.872 & 0.967 & \B{1.153} & 0.623 & 1.797 & 0.632 & 0.668 & 0.754 \\
VGGT~\citep{vggt} & 0.271 & 0.173 & 0.305 & 0.180 & 0.841 & 0.942 & 1.310 & 0.763 & 1.932 & 1.018 & 0.666 & 0.751 \\
VGGT-$\Omega$~\citep{vggtomega} & \B{0.061} & \B{0.038} & \B{0.056} & \B{0.036} & \B{0.906} & \B{0.984} & \U{1.207} & 0.624 & 1.921 & 0.628 & 0.670 & 0.758 \\
DAGE~\citep{dage} & 0.207 & 0.140 & 0.213 & 0.131 & 0.858 & 0.968 & 1.305 & \U{0.619} & 1.687 & \U{0.485} & 0.645 & 0.714 \\
SegVGGT~\citep{segvggt} & 0.299 & 0.210 & 0.306 & 0.185 & 0.839 & 0.949 & 2.334 & 1.365 & 1.976 & 0.831 & 0.685 & 0.774 \\
IGGT~\citep{iggt} & 0.254 & 0.164 & 0.270 & 0.154 & 0.866 & 0.966 & 1.455 & 0.830 & 1.953 & 1.036 & \U{0.687} & \U{0.780} \\
IGGT4D~\citep{iggt4d} & 0.253 & 0.186 & 0.296 & 0.174 & 0.853 & 0.926 & 2.041 & 1.062 & \B{1.544} & 0.678 & \B{0.691} & \B{0.784} \\
\textbf{Ours} & \U{0.076} & \U{0.041} & \U{0.070} & \U{0.037} & \U{0.893} & \U{0.981} & 1.253 & \B{0.600} & \U{1.561} & \B{0.470} & 0.648 & 0.719 \\
\bottomrule
\end{tabular}}
\end{table*}

\begin{table*}[t]
\centering
\caption{\textbf{Point map estimation on 7-Scenes~\citep{sevenscenes} and NRGBD~\citep{nrgbd}.} Metrics are the same as in Tab.~\ref{tab:pointmap-eth-dtu}. Keyframes are selected every 200 (sparse) and 40 (dense) frames for 7-Scenes, and every 500 (sparse) and 100 (dense) frames for NRGBD.}
\label{tab:pointmap}
\setlength{\tabcolsep}{2.4pt}
\renewcommand{\arraystretch}{1.05}
\resizebox{\textwidth}{!}{%
\begin{tabular}{l|cc cc cc|cc cc cc}
\toprule
& \multicolumn{6}{c|}{\textbf{7-Scenes}} & \multicolumn{6}{c}{\textbf{NRGBD}} \\
\cmidrule(lr){2-7} \cmidrule(lr){8-13}
\textbf{Method}
& \multicolumn{2}{c}{Acc. $\downarrow$} & \multicolumn{2}{c}{Comp. $\downarrow$} & \multicolumn{2}{c|}{NC. $\uparrow$}
& \multicolumn{2}{c}{Acc. $\downarrow$} & \multicolumn{2}{c}{Comp. $\downarrow$} & \multicolumn{2}{c}{NC. $\uparrow$} \\
\cmidrule(lr){2-3} \cmidrule(lr){4-5} \cmidrule(lr){6-7}
\cmidrule(lr){8-9} \cmidrule(lr){10-11} \cmidrule(lr){12-13}
& Mean & Med. & Mean & Med. & Mean & Med.
& Mean & Med. & Mean & Med. & Mean & Med. \\
\midrule
\multicolumn{13}{l}{\textit{Sparse View}} \\
DA3~\citep{da3} & 0.041 & 0.030 & 0.060 & 0.044 & \U{0.764} & \U{0.877} & 0.027 & \U{0.014} & \U{0.028} & \U{0.013} & \B{0.922} & \B{0.995} \\
$\pi^{3}$~\citep{pi3} & 0.047 & 0.028 & 0.074 & 0.048 & 0.741 & 0.840 & 0.024 & \U{0.014} & \U{0.028} & 0.014 & 0.909 & 0.991 \\
VGGT~\citep{vggt} & 0.044 & 0.024 & 0.056 & 0.033 & 0.733 & 0.846 & 0.050 & 0.027 & 0.067 & 0.038 & 0.881 & 0.979 \\
VGGT-$\Omega$~\citep{vggtomega} & \U{0.026} & \U{0.016} & \U{0.031} & \U{0.020} & \B{0.773} & \B{0.891} & \B{0.021} & \B{0.012} & \B{0.018} & \B{0.011} & \U{0.911} & 0.992 \\
DAGE~\citep{dage} & 0.042 & 0.023 & 0.058 & 0.036 & 0.675 & 0.747 & 0.035 & 0.017 & 0.037 & 0.015 & 0.882 & 0.991 \\
SegVGGT~\citep{segvggt} & 0.048 & 0.031 & 0.063 & 0.039 & 0.741 & 0.850 & 0.046 & 0.022 & 0.051 & 0.024 & 0.878 & 0.988 \\
IGGT~\citep{iggt} & 0.045 & 0.030 & 0.062 & 0.040 & 0.752 & 0.866 & 0.038 & 0.020 & 0.046 & 0.020 & 0.910 & \U{0.993} \\
IGGT4D~\citep{iggt4d} & 0.037 & 0.024 & 0.064 & 0.047 & 0.760 & 0.871 & 0.028 & 0.017 & 0.031 & 0.014 & \B{0.922} & \B{0.995} \\
\textbf{Ours} & \B{0.021} & \B{0.012} & \B{0.028} & \B{0.016} & 0.708 & 0.794 & \U{0.023} & \B{0.012} & \B{0.018} & \B{0.011} & 0.895 & 0.989 \\
\midrule
\addlinespace
\multicolumn{13}{l}{\textit{Dense View}} \\
DA3~\citep{da3} & 0.018 & 0.007 & 0.023 & 0.009 & \U{0.689} & \B{0.795} & \U{0.013} & \B{0.006} & 0.014 & 0.006 & \U{0.890} & 0.987 \\
$\pi^{3}$~\citep{pi3} & 0.016 & 0.007 & 0.022 & 0.011 & \U{0.689} & 0.792 & \U{0.013} & \U{0.007} & 0.014 & 0.006 & 0.874 & 0.981 \\
VGGT~\citep{vggt} & 0.022 & 0.008 & 0.026 & 0.012 & 0.666 & 0.760 & 0.015 & 0.008 & 0.015 & 0.006 & 0.871 & 0.982 \\
VGGT-$\Omega$~\citep{vggtomega} & 0.016 & 0.007 & \B{0.018} & \U{0.007} & \B{0.691} & \B{0.795} & \B{0.012} & \B{0.006} & \U{0.008} & \U{0.005} & 0.860 & 0.979 \\
DAGE~\citep{dage} & \U{0.015} & \U{0.006} & \U{0.020} & 0.009 & 0.638 & 0.709 & 0.015 & 0.008 & 0.014 & \U{0.005} & 0.842 & 0.976 \\
SegVGGT~\citep{segvggt} & 0.016 & 0.007 & 0.027 & 0.011 & 0.681 & 0.780 & 0.032 & 0.014 & 0.027 & 0.010 & 0.838 & 0.973 \\
IGGT~\citep{iggt} & 0.016 & 0.007 & 0.024 & 0.009 & 0.687 & 0.791 & 0.016 & 0.010 & 0.014 & 0.006 & \U{0.890} & \U{0.990} \\
IGGT4D~\citep{iggt4d} & 0.017 & 0.007 & 0.025 & 0.011 & 0.688 & \U{0.793} & \U{0.013} & 0.008 & 0.013 & \U{0.005} & \B{0.905} & \B{0.991} \\
\textbf{Ours} & \B{0.012} & \B{0.005} & \B{0.018} & \B{0.006} & 0.645 & 0.717 & \B{0.012} & \B{0.006} & \B{0.006} & \B{0.004} & 0.848 & 0.973 \\
\bottomrule
\end{tabular}}
\end{table*}

\subsection{Method Details}
\label{sec:suppl_model_training}
\textbf{Geometry Loss.}~The geometry loss consists of point, depth, and camera terms: $\mathcal{L}_{\mathrm{geo}}=\lambda_{\mathrm{point}}\mathcal{L}_{\mathrm{point}}+\lambda_{\mathrm{depth}}\mathcal{L}_{\mathrm{depth}}+\lambda_{\mathrm{cam}}\mathcal{L}_{\mathrm{cam}}$. We apply $\pi^{3}$-style~\citep{pi3} scale-invariant local geometry reconstruction loss and affine-invariant camera pose loss. For this purpose, we remove the designated reference camera view and supervise relative poses together with scale-aligned local point maps, making the geometry objective reference-free. Concretely, we unproject the predicted depths with the predicted cameras to obtain per-frame local point maps, estimate a scene-level scale that aligns them to the ground truth, and compute the point, depth and relative camera pose losses after scale alignment. $\mathcal{L}_{\mathrm{point}}$ and $\mathcal{L}_{\mathrm{cam}}$ are computed same as in $\pi^{3}$, while $\mathcal{L}_{\mathrm{depth}}$ follows VGGT~\citep{vggt} setting, containing aleatoric uncertainty and a gradient consistency term.

\textbf{Implementation.}~We train Entity3R on EntityWorld dataset with AdamW~\citep{adamw} optimizer for 160K iterations, using 32 NVIDIA H20-96GB GPUs. The learning rate follows a linear warm-up over 5\% of training and cosine decay for the remaining 95\%. We set a peaking learning rate of $1{\times}10^{-5}$ for the HR DINOv3~\citep{dinov3} encoder and the LR VGGT-$\Omega$~\citep{vggtomega} geometry transformer, and $1{\times}10^{-4}$ for the remaining modules. Training proceeds in two phases: the first 80K iterations train with only the geometry loss $\mathcal{L}_{\mathrm{geo}}$, and the remaining 80K iterations jointly optimize with $\mathcal{L}_{\mathrm{geo}}+\mathcal{L}_{\mathrm{ent}}$. Following VGGT~\citep{vggt}, we randomly apply color jitter, Gaussian blur, and grayscale to the input frames as data augmentation. For training stability, we normalize the ground-truth cameras, depths, and local points using the global point cloud scale. During training, HR input images are resized to a width of 1024 pixels with a random aspect ratio sampled from $[0.33, 1.0]$, and the LR branch receives frames downsampled by $1/2$ to a width of 512. We sample 1-12 frames per sequence and use 12 images per GPU. To keep training efficient, we use bf16 mixed precision, gradient checkpointing, and gradient accumulation with 2 steps. For the loss weights, we set $\lambda_{\mathrm{point}}{=}1$, $\lambda_{\mathrm{depth}}{=}1$, $\lambda_{\mathrm{cam}}{=}5$, $\lambda_{\mathrm{cls}}{=}1$ and $\lambda_{\mathrm{bce}}{=}\lambda_{\mathrm{dice}}{=}2$.

\textbf{Cross-chunk association in chunk-wise object decoding.}~We associate objects in current chunk with the historical object pool using mask-pooled instance features. Each object’s feature center is obtained by pooling the instance features within its predicted masklet and applying L2 normalization. Association follows greedy one-to-one matching with a cosine-similarity threshold of 0.9. At each step, we select the most similar pair of current and historical objects. If their similarity is no smaller than the threshold, we merge the current object into the corresponding historical object and exclude both from further matching within the current chunk. This process continues until the highest remaining similarity falls below the threshold. Each unmatched current object is then assigned a new identity. For each matched pair, we update the historical feature center using an area-weighted average followed by L2 normalization: $f \leftarrow \mathrm{L2Norm}((A_p f_p + A_c f_c)/(A_p + A_c))$, with the accumulated area updated as $A \leftarrow A_p + A_c$. Here, $f_p$ and $A_p$ denote the historical feature center and accumulated mask area, while $f_c$ and $A_c$ denote the current object’s feature center and mask area.

\textbf{Clarification of  the training/testing splits.}~For datasets that are used both for training and for evaluation, we follow the official training/testing splits provided by each dataset when such splits are officially released. When a dataset (e.g, Co3Dv2) does not provide an official split, we adopt the evaluation protocol used in VGGT to define the splits, so that our results are fairly comparable to existing baselines. In all cases, the training and evaluation sets are disjoint at the sequence level by construction of these splits. The isolation is guaranteed by the splits themselves. Moreover, we do not apply an additional deduplication step in the data preprocessing stage. We have added a clarification of this protocol to the revised manuscript.

\begin{table}[t]
  \centering
  \caption{\textbf{Dataset composition in EntityWorld.} \emph{Provided} means the dataset provides entity mask labels; \emph{Annotated} means we annotate the masks using our CropFormer+SAM2 pipeline.}
  \label{tab:entityworld_sources}
  \small
  \setlength{\tabcolsep}{2.5pt}
  \resizebox{\linewidth}{!}{%
  \begin{tabular}{lccccccc}
    \toprule
    \textbf{Dataset} & Scene Type & Dynamic & Real & Entity Masks & \#~Frames & \#~Seqs & Sample Prob. \\
    \midrule
    Mapillary~\citep{mapillary} & Outdoor-urban & Dynamic & Real & Annotated & 34K & 580 & 0.04 \\
    Waymo~\citep{waymo} & Outdoor-urban & Dynamic & Real & Annotated & 703K & 9.72K & 0.06 \\
    VirtualKITTI2~\citep{virtualkitti2} & Outdoor-urban & Dynamic & Synthetic & Provided & 42.5K & 100 & 0.04 \\
    MatrixCity~\citep{matrixcity} & Outdoor-urban & Dynamic & Synthetic & Annotated & 516K & 5.18K & 0.04 \\
    ScanNet++~\citep{scannetpp} & Indoor & Static & Real & Annotated & 1.7M & 14.9K & 0.07 \\
    HOI4D~\citep{hoi4d} & Indoor & Dynamic & Real & Annotated & 891K & 2.95K & 0.04 \\
    Hypersim~\citep{hypersim} & Indoor & Static & Synthetic & Provided & 66.8K & 690 & 0.04 \\
    Aria Synthetic Env.~\citep{aria} & Indoor & Static & Synthetic & Provided & 58.1M & 100K & 0.07 \\
    PointOdyssey~\citep{pointodyssey} & Indoor & Dynamic & Synthetic & Provided & 106.5K & 358 & 0.04 \\
    Dynamic Replica~\citep{dynamicreplica} & Indoor & Dynamic & Synthetic & Annotated & 300.6K & 1K & 0.06 \\
    Infinigen~\citep{infinigen} & Indoor & Static & Synthetic & Provided & 146K & 1.46K & 0.04 \\
    WildRGBD~\citep{wildrgbd} & Object-centric & Static & Real & Annotated & 1.41M & 7.84K & 0.05 \\
    Co3Dv2~\citep{co3dv2} & Object-centric & Static & Real & Annotated & 2.65M & 32.8K & 0.06 \\
    Kubric~\citep{kubric} & Object-centric & Dynamic & Synthetic & Provided & 135.1K & 5.63K & 0.04\\
    MapFree~\citep{mapfree} & Mixed & Dynamic & Real & Annotated & 504.4K & 2.78K & 0.04 \\
    DL3DV~\citep{dl3dv} & Mixed & Static & Real & Annotated & 2.06M & 19.4K & 0.07 \\
    TartanAir~\citep{tartanair} & Mixed & Dynamic & Synthetic & Provided & 306.6K & 369 & 0.04 \\
    OmniWorld-Game~\citep{omniworld} & Mixed & Dynamic & Synthetic & Annotated & 813.5K & 11.5K & 0.06 \\
    EntityWorld-Game & Mixed & Dynamic & Synthetic & Annotated & 14.3M & 143K & 0.10 \\
    \midrule
    \textbf{EntityWorld} & -- & -- & -- & -- & \textbf{84.8M} & \textbf{360K} & \textbf{1.00}\\
    \bottomrule
  \end{tabular}%
  }
\end{table}

\begin{figure}[t]
    \centering
    \includegraphics[width=0.9\linewidth]{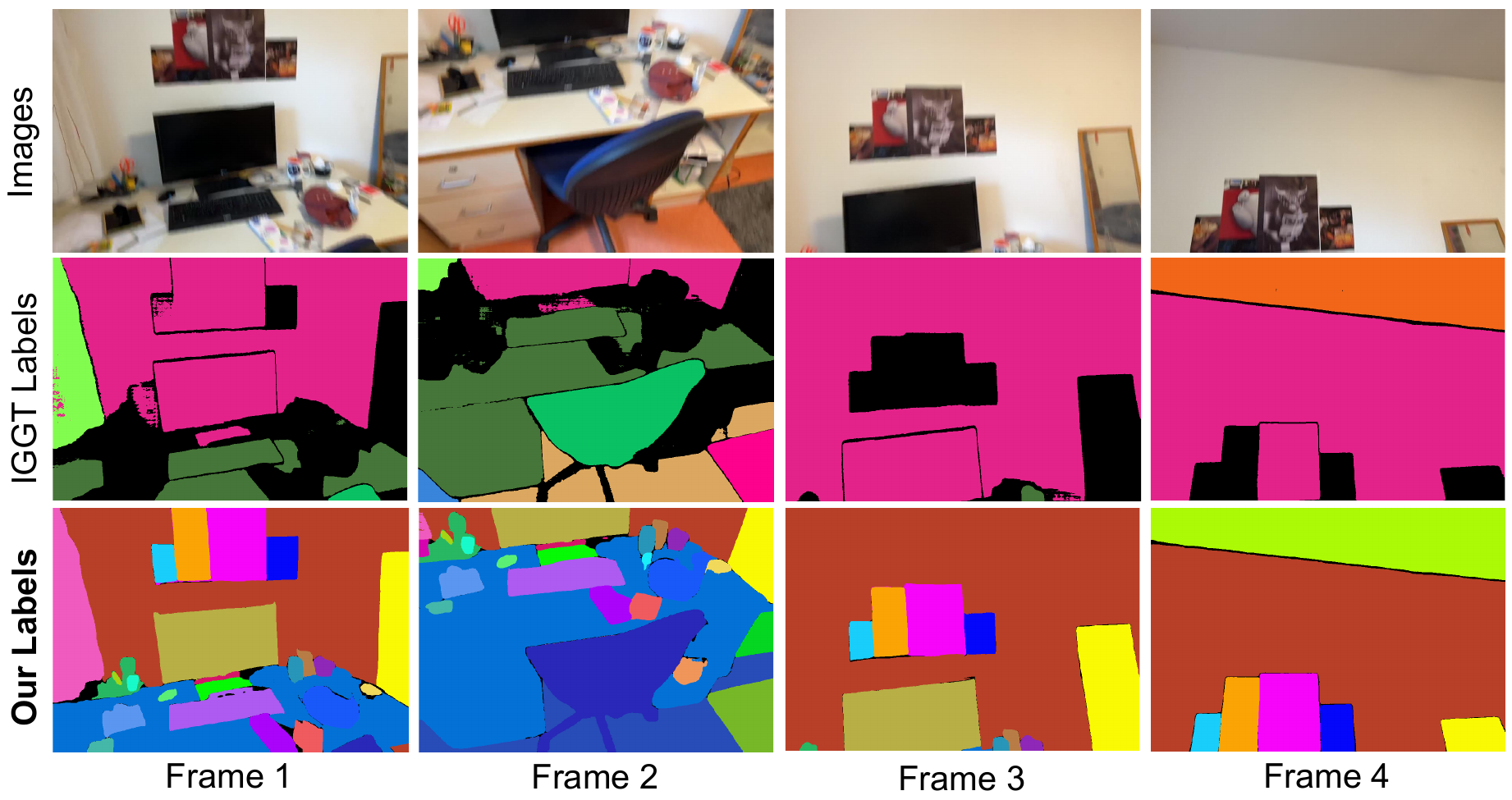}
    \caption{Comparison of entity masklet annotations. Compared with the IGGT~\citep{iggt} labels, our annotation pipeline generates more complete and higher-quality entity masklets.}
    \label{fig:suppl_label_cmp}
\end{figure}

\begin{table*}[t]
    \centering
    \caption{\textbf{Quantitative annotation assessment} on our video entity segmentation benchmarks.}
    \vspace{-2mm}
    \label{tab:annotation_assessment}
    \small
    \setlength{\tabcolsep}{4.5pt}
    \renewcommand{\arraystretch}{1.05}
    \resizebox{0.9\textwidth}{!}{%
    \begin{tabular}{l c cc cc cc}
    \toprule
    \multirow{2}{*}[-0.4em]{\textbf{Method}}
    & \textbf{DAVIS}
    & \multicolumn{2}{c}{\textbf{VIPSeg}}
    & \multicolumn{2}{c}{\textbf{Hypersim}}
    & \multicolumn{2}{c}{\textbf{Waymo}} \\
    \cmidrule(lr){2-2} \cmidrule(lr){3-4} \cmidrule(lr){5-6} \cmidrule(lr){7-8}
    & $\mathcal{J}\&\mathcal{F}$ $\uparrow$
    & VEQ $\uparrow$ & $\mathrm{STQ}_{\mathrm{EN}}$ $\uparrow$
    & VEQ $\uparrow$ & $\mathrm{STQ}_{\mathrm{EN}}$ $\uparrow$
    & VEQ $\uparrow$ & $\mathrm{STQ}_{\mathrm{EN}}$ $\uparrow$ \\
    \midrule
    IGGT annotations
    & 67.4 & 63.4 & 49.0 & 32.0 & 27.9 & 38.3 & 24.5 \\
    Our annotations
    & \textbf{79.7} & \textbf{72.2} & \textbf{54.2} & \textbf{34.5} & \textbf{29.4} & \textbf{40.4} & \textbf{25.4} \\
    \bottomrule
    \end{tabular}}
    \end{table*}

\subsection{EntityWorld Dataset Details}
\label{sec:suppl_entityworld_composition}
We first collect several publicly available geometric datasets in the community: Mapillary~\citep{mapillary}, Waymo~\citep{waymo}, VirtualKITTI2~\citep{virtualkitti2}, MatrixCity~\citep{matrixcity}, ScanNet++~\citep{scannetpp}, HOI4D~\citep{hoi4d}, Hypersim~\citep{hypersim}, Aria Synthetic Environments~\citep{aria}, PointOdyssey~\citep{pointodyssey}, Dynamic Replica~\citep{dynamicreplica}, Infinigen~\citep{infinigen}, WildRGBD~\citep{wildrgbd}, Co3Dv2~\citep{co3dv2}, Kubric~\citep{kubric}, MapFree~\citep{mapfree}, DL3DV~\citep{dl3dv}, TartanAir~\citep{tartanair}, and OmniWorld-Game~\citep{omniworld}. We further add our newly collected EntityWorld-Game sequences. Tab.~\ref{tab:entityworld_sources} lists every source used to build EntityWorld, together with its scene type, motion type, capture type, entity-mask origin, the number of frames and sequences, and the sampling probabilities. In total, EntityWorld contains nearly 84.8M frames and 360K sequences, spanning diverse domains including indoor/ourdoor and dynamic/static scenes. We provide some data samples of EntityWorld in Fig.~\ref{fig:suppl_data_sample}. Moreover, we quantitatively evaluate annotation quality by treating the labeling pipelines themselves as predictors. We run both the IGGT labeling pipeline and our pipeline on the video entity segmentation benchmarks used in our work, and report the resulting metrics in Tab.~\ref{tab:annotation_assessment}. Across all tracks, our annotations achieve higher scores than IGGT labels, which quantitatively supports the qualitative comparisons in Fig. 7.

\begin{figure}[t]
    \centering
    \includegraphics[width=0.9\linewidth]{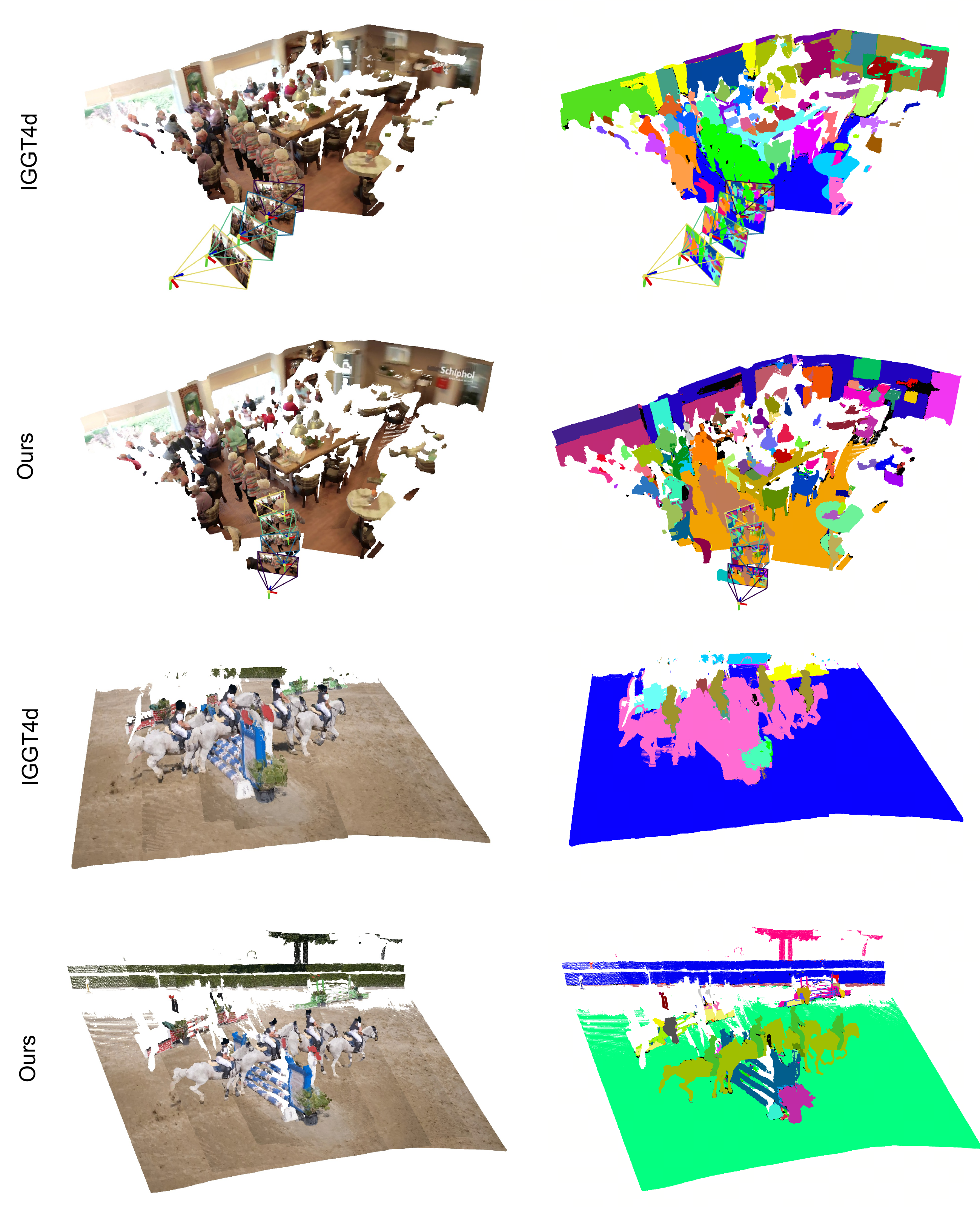}
    \caption{Additional qualitative comparisons of 3D entity segmentation.}
    \label{fig:suppl_qual}
\end{figure}

\begin{figure}[t]
    \centering
    \vspace{-3mm}
    \includegraphics[width=0.95\linewidth]{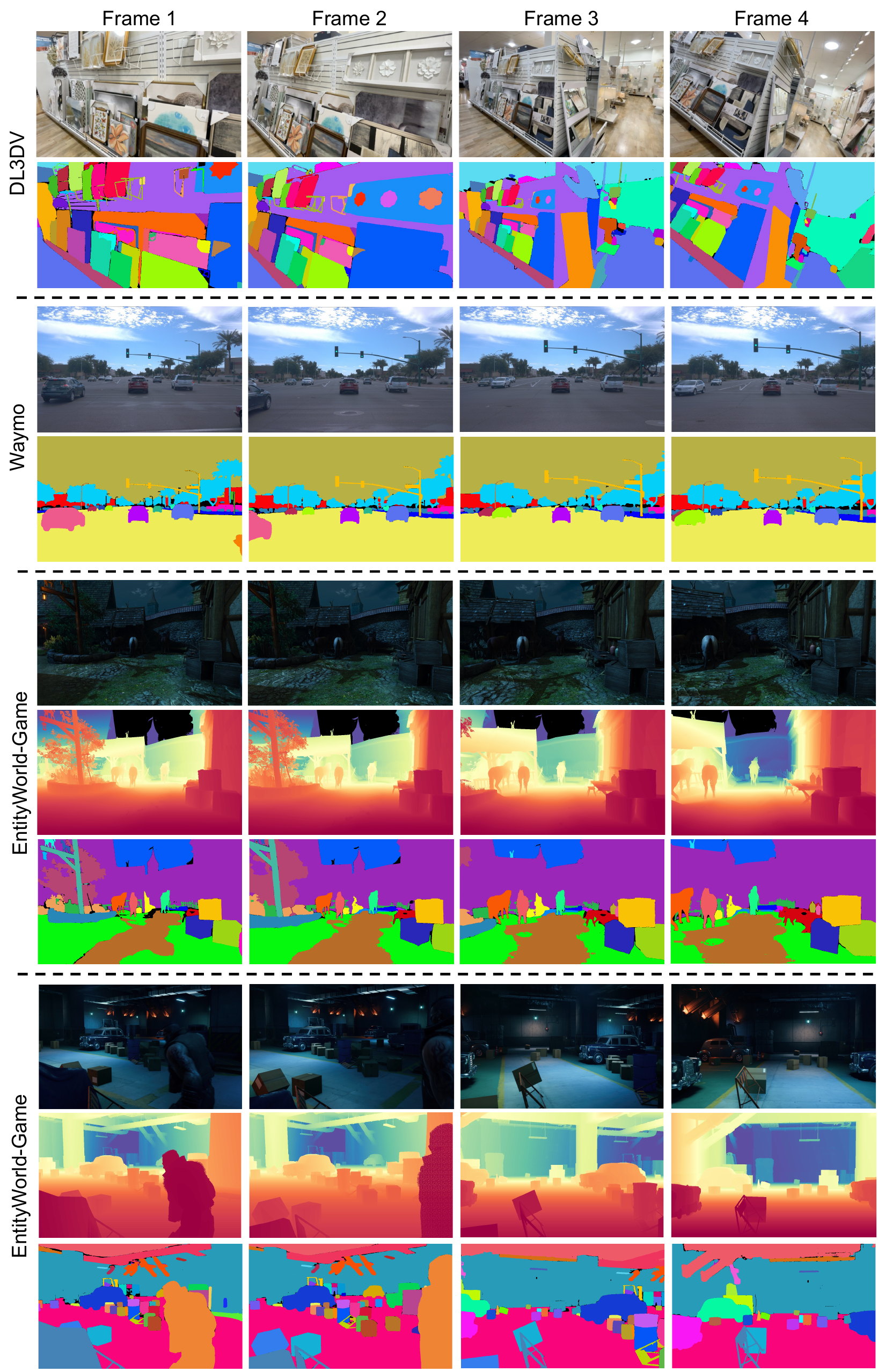}
    \vspace{-2mm}
    \caption{EntityWorld data samples across diverse domains. For each sequence we show RGB frames together with the corresponding entity masklets. For our self-collected EntityWorld-Game sequences, we additionally show the rendered depth maps.}
    \label{fig:suppl_data_sample}
\end{figure}

\subsection{Additional Experimental Results}
\label{sec:suppl_qualitative}
\textbf{Qualitative results.}~Here we provide more qualitative comparison results in Fig.~\ref{fig:suppl_qual}.

\textbf{Training other methods on EntityWorld.}~We retrain IGGT4D~\citep{iggt4d} on EntityWorld under a controlled setting. Compared with the original IGGT4D, EntityWorld training improves the performance, as listed in Tab.~\ref{tab:iggt4d_entityworld}, demonstrating that EntityWorld is useful beyond our own model and benefits other methods as well.

\begin{table*}[t]
    \centering
    \caption{\textbf{Evaluation results of retraining IGGT4D on EntityWorld.}}
    \vspace{-2mm}
    \label{tab:iggt4d_entityworld}
    \small
    \setlength{\tabcolsep}{4.5pt}
    \renewcommand{\arraystretch}{1.05}
    \resizebox{\textwidth}{!}{%
    \begin{tabular}{l | cccc | cccc}
    \toprule
    \multirow{2}{*}[-0.4em]{\textbf{Method}}
    & \multicolumn{4}{c|}{\textbf{VIPSeg}~(Shorter Seqs.)}
    & \multicolumn{4}{c}{\textbf{Waymo}~(Longer Seqs.)} \\
    \cmidrule(lr){2-5} \cmidrule(lr){6-9}
    & VEQ $\uparrow$ & $\mathrm{STQ}_{\mathrm{EN}}$ $\uparrow$ & VPQ $\uparrow$ & STQ $\uparrow$
    & VEQ $\uparrow$ & $\mathrm{STQ}_{\mathrm{EN}}$ $\uparrow$ & VPQ $\uparrow$ & STQ $\uparrow$ \\
    \midrule
    IGGT4D
    & 58.1 & 48.1 & 27.8 & 35.2 & 27.1 & 17.5 & 15.2 & 10.7 \\
    IGGT4D (trained on EntityWorld)
    & \textbf{60.3} & \textbf{50.6} & \textbf{28.9} & \textbf{37.5}
    & \textbf{29.3} & \textbf{18.9} & \textbf{16.8} & \textbf{12.0} \\
    \bottomrule
    \end{tabular}}
    \end{table*}
\begin{table*}[t]
    \centering
    \caption{\textbf{Entity recall and identity association} in challenging scenarios, including sequences with more than 200 entities, similar-appearance entities, or entities re-emergencing after occlusion.}
    \vspace{-2mm}
    \label{tab:recall_association}
    \small
    \setlength{\tabcolsep}{4.5pt}
    \renewcommand{\arraystretch}{1.05}
    \resizebox{0.9\textwidth}{!}{%
    \begin{tabular}{l | cc | cc | cc}
    \toprule
    \multirow{2}{*}[-0.4em]{\textbf{Method}}
    & \multicolumn{2}{c|}{\textbf{$>$200 entities}}
    & \multicolumn{2}{c|}{\textbf{Similar appearance}}
    & \multicolumn{2}{c}{\textbf{Re-emergence}} \\
    \cmidrule(lr){2-3} \cmidrule(lr){4-5} \cmidrule(lr){6-7}
    & Recall $\uparrow$ & AQ $\uparrow$
    & Recall $\uparrow$ & AQ $\uparrow$
    & Recall $\uparrow$ & AQ $\uparrow$ \\
    \midrule
    IGGT4D
    & 0.498 & 0.156 & 0.563 & 0.179 & 0.482 & 0.108 \\
    Ours (full-context decoding)
    & 0.709 & 0.208 & 0.787 & 0.274 & 0.723 & 0.185 \\
    Ours (chunk-wise decoding, $S{=}1$)
    & \textbf{0.776} & \textbf{0.272} & \textbf{0.812} & \textbf{0.304} & \textbf{0.767} & \textbf{0.213} \\
    \bottomrule
    \end{tabular}}
    \end{table*}

\textbf{Stability of chunk-wise decoding in challenging scenarios.}~To evaluate the stability of chunk-wise object decoding on challenging sequences with more than 200 distinct entities, on re-emergence after occlusion, and on entities with similar appearance, we add class-agnostic evaluation on the Waymo test set with 100 sequences. The evaluation is conducted on three groups: (1) More than 200 entities. We use the 25 sequences whose object count exceeds 200, which contain 477 entities per sequence on average. (2) Similar appearance. We evaluate only cars and pedestrians in all 100 sequences. Cars are similar to other cars, and pedestrians are similar to other pedestrians. This gives 3,850 identities in total. (3) Re-emergence after occlusion. We evaluate every entity, in all 100 sequences, that disappears for at least one frame and later reappears. This gives 2,902 identities. We use two metrics, Recall and AQ. Recall is the fraction of ground-truth object masks that overlap a predicted mask with IoU greater than 0.5. AQ is the association quality score proposed in~\citep{weber2021step}. Tab.~\ref{tab:recall_association} shows that, in these challenging scenarios, chunk-wise decoding is better than full-context decoding on both recall and AQ, and both are better than IGGT4D. 

\textbf{Initialization with retrained VGGT-$\Omega$ checkpoint.}~In Entity3R training, we use VGGT-Omega-1B-512 checkpoint (https://github.com/facebookresearch/vggt-omega) for backbone initialization. The potential data contamination dose not affect the video entity/panoptic segmentation benchmarks, that are our core goal. To address the influence on reconstruction benchmarks, we have retrained Entity3R with the newly released VGGT-Omega-1B-416-Reproduction checkpoint. As shown in Tab.~\ref{tab:omega-416}, the retrained model remains competitive with DAGE and VGGT-omega and obtains the best result on most metrics.

\begin{table*}[t]
    \centering
    \caption{\textbf{Reconstruction metrics with initialization from retrained VGGT-$\Omega$ checkpoint.}}
    \label{tab:omega-416}
    \setlength{\tabcolsep}{3.0pt}
    \renewcommand{\arraystretch}{1.05}
    \resizebox{\textwidth}{!}{%
    \begin{tabular}{l|cc cc|cc ccc}
    \toprule
    \multirow{3}{*}[-2ex]{\textbf{Method}}
    & \multicolumn{4}{c|}{\textbf{Video Depth Estimation}}
    & \multicolumn{5}{c}{\textbf{Camera Pose Estimation}} \\
    \cmidrule(lr){2-5} \cmidrule(lr){6-10}
    & \multicolumn{2}{c}{\textbf{Bonn}} & \multicolumn{2}{c|}{\textbf{Sintel}}
    & \multicolumn{2}{c}{\textbf{RealEstate10K}}
    & \multicolumn{3}{c}{\textbf{Sintel}} \\
    \cmidrule(lr){2-3} \cmidrule(lr){4-5} \cmidrule(lr){6-7} \cmidrule(lr){8-10}
    & Abs Rel $\downarrow$ & $\delta_1$ $\uparrow$
    & Abs Rel $\downarrow$ & $\delta_1$ $\uparrow$
    & AUC/$5$ $\uparrow$ & AUC/$15$ $\uparrow$
    & ATE $\downarrow$ & RPE-t $\downarrow$ & RPE-r $\downarrow$ \\
    \midrule
    DAGE~\citep{dage}
      & \underline{0.055} & \textbf{0.978}
      & \underline{0.274} & \underline{0.665}
      & 52.49 & 74.67
      & 0.141 & 0.054 & 0.416 \\
    VGGT-$\Omega$~\citep{vggtomega} (retrained)
      & 0.113 & 0.933
      & 0.289 & 0.629
      & \underline{54.66} & \underline{77.13}
      & \underline{0.081} & \underline{0.045} & \textbf{0.340} \\
    Ours (retrained)
      & \textbf{0.052} & \underline{0.976}
      & \textbf{0.253} & \textbf{0.686}
      & \textbf{61.48} & \textbf{79.40}
      & \textbf{0.079} & \textbf{0.043} & \underline{0.347} \\
    \bottomrule
    \end{tabular}}
\end{table*}

\subsection{Runtime Analysis}
\label{sec:suppl_runtime}
Tab.~\ref{tab:runtime} reports the per-component runtime on a 100-frame sequence with an NVIDIA H20-96GB GPU.
First, our query-based object decoding costs less than 3\,s across settings, whereas IGGT4D~\citep{iggt4d} spends 20.9s / 57.3s on clustering post-processing at the two resolutions.
Second, feeding high-resolution inputs directly into the geometry transformer is prohibitively expensive: at a width of about ${\sim}$1024, IGGT4D and our LR variant (w/o HR branch) both run the geometry transformer at full resolution, taking 231.7s / 62.1\,GB and 100.8s / 76.9\,GB, respectively.
In contrast, our full Entity3R model keeps the geometry transformer on low-resolution tokens and recovers fine-grained details with a frame-wise HR encoder that bypasses global attention, reducing the cost to 18.7s with only 27.3\,GB peak memory---about $5{\times}$--$12{\times}$ faster and much more memory-efficient than pushing high-resolution inputs through the geometry transformer.
\begin{table}[t]
\centering
\caption{\textbf{Runtime breakdown} measured on a 100-frame sequence with an NVIDIA H20-96GB GPU.
Times are in seconds; Mem.\ is the peak GPU memory in GB. Geo. Trans. is the geometry transformer; HR Enc. is the HR encoder; Geo. is the geometry (camera and depth) heads; Inst. is the instance head; Cluster is the clustering post-processing; Query is our query decoding.
$^\dag$Offline inference with full global attention for IGGT4D~\citep{iggt4d}, matching Tab.~\ref{tab:entity-seg}.}
\label{tab:runtime}
\small
\setlength{\tabcolsep}{2.4pt}
\resizebox{0.9\linewidth}{!}{%
\begin{tabular}{l|c|cc|cc|cc|c c}
\toprule
\multirow{2}{*}[-0.4em]{\textbf{Method}}
& \multirow{2}{*}[-0.4em]{\shortstack{Input\\Width}}
& \multicolumn{2}{c|}{\textbf{Backbone}}
& \multicolumn{2}{c|}{\textbf{Head}}
& \multicolumn{2}{c|}{\textbf{Object Decoding}}
& \multirow{2}{*}[-0.4em]{Total}
& \multirow{2}{*}[-0.4em]{Mem.} \\
\cmidrule(lr){3-4} \cmidrule(lr){5-6} \cmidrule(lr){7-8}
&
& Geo.\ Trans. & HR Enc.
& Geo & Inst.
& Cluster & Query
& & \\
\midrule
IGGT4D~\citep{iggt4d}$^\dag$
& 504
& 12.3 & --
& 1.3 & 1.7
& 20.9 & --
& 36.2 & 21.0 \\
Ours (w/o HR branch)
& 512
& 7.6 & --
& 0.3 & 1.2
& -- & 0.6
& \B{9.7} & 22.6 \\
\midrule
IGGT4D~\citep{iggt4d}$^\dag$
& 1008
& 162.1 & --
& 5.3 & 7.1
& 57.3 & --
& 231.8 & 62.1 \\
Ours (w/o HR branch)
& 1024
& 94.3 & --
& 1.0 & 4.0
& -- & 2.4
& 101.7 & 76.9 \\
Ours-HR
& 1024
& 7.6 & 5.9
& 0.9 & 3.9
& -- & 1.2
& \U{19.5} & 27.3 \\
\bottomrule
\end{tabular}}
\end{table}

\subsection{Limitations}
Despite strong entity segmentation and competitive geometry, Entity3R still has several limitations.
First, although chunk-wise object decoding improves inference scalability, the geometry transformer still consumes all frames jointly, so very long videos remain constrained by backbone memory rather than by the query decoder. Extending Entity3R to extremely long sequences is a promising direction, and we can draw on recent streaming geometric reconstruction methods~\citep{lingbotmap,streamvggt,infinitevggt} to scale the backbone beyond full-sequence attention.
Second, our automatic CropFormer$+$SAM2~\citep{cropformer,sam2} annotation pipeline tends to under-segment \emph{stuff} regions, e.g., different buildings are often merged into a single entity. We hope to construct high-quality synthetic datasets with instance-level stuff masks that distinguish such regions, and to fine-tune Entity3R on them to better separate stuff entities.

\end{document}